\documentclass{ieeeaccess}
\usepackage{cite}
\usepackage{amsmath,amssymb,amsfonts}
\usepackage{algorithmic}
\usepackage{graphicx}
\usepackage{textcomp}
\DeclareUnicodeCharacter{00E1}{\'{a}}
\usepackage[utf8]{inputenc}
\usepackage{bm}
\makeatletter
\AtBeginDocument{\DeclareMathVersion{bold}
\DeclareUnicodeCharacter{00A0}{\nobreakspace}
\DeclareUnicodeCharacter{2009}{\,} 
\DeclareUnicodeCharacter{2217}{\,} 
\DeclareUnicodeCharacter{FFFD}{\,} 
\SetSymbolFont{operators}{bold}{T1}{times}{b}{n}
\SetSymbolFont{NewLetters}{bold}{T1}{times}{b}{it}
\SetMathAlphabet{\mathrm}{bold}{T1}{times}{b}{n}
\SetMathAlphabet{\mathit}{bold}{T1}{times}{b}{it}
\SetMathAlphabet{\mathbf}{bold}{T1}{times}{b}{n}
\SetMathAlphabet{\mathtt}{bold}{OT1}{pcr}{b}{n}
\SetSymbolFont{symbols}{bold}{OMS}{cmsy}{b}{n}
\renewcommand\boldmath{\@nomath\boldmath\mathversion{bold}}}
\makeatother

\def\BibTeX{{\rm B\kern-.05em{\sc i\kern-.025em b}\kern-.08em
    T\kern-.1667em\lower.7ex\hbox{E}\kern-.125emX}}

\ifCLASSINFOpdf
\else
\fi
\usepackage{graphicx}
\usepackage{float}
\usepackage{amsmath}

\begin{document}
%
\title{Safe Pathfinding in BIM Worlds Utilizing Dynamic MHA* Algorithms using APF and Natural Language Processing}
%
%
%
\history{Date of publication xxxx 00, 0000, date of current version xxxx 00, 0000.}
\doi{10.1109/ACCESS.2024.0429000}

\title{Safe and Trustworthy Robot Pathfinding with BIM, MHA*, and NLP}
\author{\uppercase{Mani Amani}\authorrefmark{1,2},
\uppercase{Reza Akhavian}\authorrefmark{1}}

\address[1]{Department of Civil, Construction, and Environmental Engineering, San Diego State University, San Diego, CA, 92182 }
\address[2]{Department of Electrical and Computer Engineering, University of California, San Diego, CA, 92092)}

\tfootnote{This work was supported in part by the National Science Foundation (NSF) under Grant 2047138 and a scholarship funded by Grant  1930546.}

\corresp{Corresponding author: Reza Akhavian (e-mail:rakhavian@sdsu.edu).}

\markboth{September 2024}%
{{Amani \& Akhavian}: Safe and Trustworthy Robot Pathfinding with BIM, MHA*, and NLP}

\begin{abstract}


Construction robotics has gained significant traction in research and development, yet deploying robots in construction environments presents unique challenges. Construction sites are characterized by dynamic conditions, domain-specific tasks, and complex navigation requirements that make traditional robot pathfinding approaches insufficient. While methods like simultaneous localization and mapping (SLAM) offer viable solutions for robot navigation, they often require considerable computational resources due to their sensor precision demands and data processing needs. This paper presents a novel approach that leverages building information modeling (BIM) for efficient, domain-specific pathfinding by utilizing both spatial and semantic information. We integrate a multi-heuristic A* (MHA*) algorithm with artificial potential fields (APF) derived from BIM spatial data, while employing large language models (LLMs) to process BIM's textual information for dynamic obstacle avoidance. Our experimental results demonstrate an 80\% improvement in robot-obstacle clearance while maintaining comparable path lengths to traditional methods. This approach provides a computationally efficient solution for safe robot navigation in complex construction environments.

\end{abstract}


\maketitle

%
\section{Introduction}
Robotic pathfinding has been an active area of research since the 1960s. Pathfinding deals with the problem of how to move a robot from one point to another. Many algorithms have been developed to tackle autonomous agents' pathfinding and guidance for known and unknown environments \cite{MDPIdronesPath}. In known environments, approaches such as artificial potential fields (APF) generate attractive and repulsive potential forces that guide the agent through the global environment \cite{APFOG}. Other approaches including heuristic search and graph search algorithms (e.g., A* and Dijkstra) aim to find the shortest path from the start goal to the end goal \cite{RefA2} \cite{RefA1}. However, in some scenarios, the shortest path is not the most optimal path. This is particularly the case in the construction industry domain. Given the dynamic nature of the construction sites, collisions can occur when rapid environmental changes transform a known environment into an unknown one. Therefore, It is extremely important to plan for this dynamism using object avoidance approaches both in real-time and in the mission planning stage. 
\par

Real-time collision avoidance methods have been developed using visual information \cite{VisualObjectAvoidance} and LiDAR-based approaches \cite{MDPILIDar}. However, these methods can increase the computational and power costs of the agent. Building information modeling (BIM), which results in a digital 3D representation of a facility's physical and functional characteristics,  provides a rich source of spatio-temporal and semantic information that can be leveraged with very low computational cost. BIM has shown significant promise in the field of construction robotics. For example, BIM proved beneficial in simultaneous localization and mapping (SLAM) in construction research where researchers address building progress monitoring using robots and mapping techniques \cite{ASCEBIMLOCALization}. Additionally, the presence of textual information embedded in BIM 3D models allows leveraging natural language processing (NLP) for explainability, thus improving the pathfinding process. Through the use of large language models (LLMs), natural language has been successfully used as a tool for robotic guidance as well \cite{NikolayLLMA}. Natural language not only helps with enhancing human-robot interaction and semantic understanding of the environment and task objectives, but it also enables explainability, a pivotal feature for the widespread integration of robots in human-centered industrial settings that has been shown to have a significant impact on robot trustworthiness in construction \cite{EMAMINEJAD2022104298}\cite{Explainability}. In this paper, we propose to use the textual BIM family properties to detect the uncertainty of accurate mapping from the semantic information in the BIM models and use it to scale and generate the APFs depending on the objects' risks dynamically. Additionally, we generate explanations on decisions made by the algorithm to enhance trust and communications. The values from the APFs are then formulated into heuristics to be used with an admissible heuristic such as Euclidean distance, which results in a safe and aware pathfinding.

\section{Literature Review}
\subsection{BIM localization}
BIM serves as a comprehensive source of data for learning-based methods such as machine learning (ML) algorithms \cite{zabin2022applications}. The robust and diverse semantic information embedded in BIM models can assist in developing new ML models for more efficient, safer, and more sustainable construction processes. That is why BIM has been leveraged  in important robotics applications\cite{zhang2022towards}, aiding in such tasks as localization and mapping \cite{moura2021bim}
and pathfinding \cite{zhou2020accurate}. However, due to the discrepancies between as-designed and as-built models (that is the difference between what is modeled and what is built in the field), relying only on BIM or A* (without using methods such as APF) can result in collisions in real-time deployments. Recent studies have managed to achieve a localization error of 20cm \cite{BIMLocalationzerpoint2} which is not considered safe around humans and in jobsite environments.
\par

Indoor localization of workers is a well-researched topic \cite{liu2017scene}. Previous works have been able to calibrate and match objects between objects and virtual environments \cite{schmidt2014automatic}. Methods such as passive
radio frequency identification (RFID) and wireless \cite{costin2015fusing, chen2014integration}, visual and inertial sensors \cite{yilmaz2016indoor} and monocular cameras \cite{deng2017bim} have shown advancements in the localization of objects and workers in a BIM world. However, all of these measurements have innate errors which can impose risks in mobile robot path planning. To use BIM models for robot path planning, these discrepancies and errors need to be accounted for to ensure safe human-robot collaboration and risk mitigation.
\par
\subsection{A* Search}
A* search algorithm is a graph traversal and pathfinding algorithm that is used in robotics and computer science \cite{russell2016artificial}. Outside of the field of robotics, many other fields also use the A* algorithm to solve their graph traversal problems. In video games, A* is an extremely popular algorithm for agent pathfinding \cite{Videogame}. Many game engines such as Unity have their own in-built A* implementation for developers to use. The algorithm uses weighted graphs to find the shortest path between the start node and the goal node, and a heuristic function calculates the cost of each node. As long as it is admissible, meaning it never overestimates the cost, the algorithm is guaranteed to generate the shortest path with efficient computational requirements. This is incredibly useful in known environments, and many current studies focus on representing real-world environments in a virtual setting using digital twins for path planning \cite{halder2024robotic}\cite{TWIN} \cite{TwinPath}. Recent advancements in digital twins have increased the match between the virtual and the real world, but certain discrepancies are inevitable. Since the original A* algorithm assumes the environment is fully known, these discrepancies can introduce risks to the system. With the introduction of the A* algorithm, a multitude of new variations to improve pathfinding and computational efficiency have emerged. However, one of A*'s biggest drawbacks is the worst-case memory and runtime needs that grow exponentially with respect to the depth of the search \cite{WeightedAstar}. Many alternatives have been developed such as the weighted A* \cite{WeightedAstar}, anytime A* variants \cite{Anytime}, and multi-resolution variants \cite{du2020multi}. 
\subsection{Artificial Potential Fields}
APF is a well-established path-finding method used in robotics \cite{oldAPF}. The main idea of APFs is to generate a potential field with a combination of attractive and repulsive forces. Obstacles generate repulsive behavior and the goal will be represented as an attractive force that pulls the agent towards itself. APFs have proven very useful in robot path planning. However, some inherent issues such as (1) the absence of a feasible path in dense
obstacle spaces; (2) the path trajectory goes beyond the equilibrium position, oscillating, or repeatedly closed-loop in the narrow
space; and (3) it is trapped in the local minima before reaching the target, introduce functional drawbacks to real-world applications \cite{wu2023robot}. Previously, APF has been used with rapid tree search algorithms \cite{wu2022apf}. Given that BIM and digital twins have detailed 3D representations, APF is a promising candidate to be incorporated with BIM for planning. Hence, we propose to use the APF as a potentially inadmissible heuristic given by the rich but naturally error-prone spatial data of BIM. Previous studies have investigated the use of A* and APF together for local and global path planning for multiple surface vehicle formations \cite{sang2021hybrid}. For example, Zhang et al. have used APFs with A* improved AGV pathfinding \cite{zhang2024agv}. However, a major limitation of those works is that the potential value of the cost function is incorporated linearly, formulating a new heuristic function. The APF is not necessarily an admissible heuristic and the admissibility of a new heuristic was not addressed in that work either. That is why we propose to use the multi-heuristic A* (MHA*) that can use multiple inadmissible heuristics (in this case the potential values from the APF), and a consistent and admissible heuristic (Euclidean distance) to provide optimality and completeness that is necessary for A*'s success \cite{MHA*}.
\subsection{Trust and Reasoning Capabilities in LLMs}
Lack of transparency in construction robotics has been identified as one of the main barriers to the widespread adoption of industrial and collaborative robots \cite{TrustNewsha}. Many previous works have identified that robot communication through natural language can increase trust between the human and the artificial agent \cite{EricTrustGPT} \cite{BagheriTrustExplain}.  LLMs have shown advanced capabilities in logical deduction and reasoning \cite{creswell2022faithfulreasoningusinglarge}.  Furthermore, LLMs have been able to utilize multi-modal architectures to be able to extend their reasoning capabilities to multiple modalities such as images and videos creating Visual Language Models (VLM) \cite{alayrac2022flamingovisuallanguagemodel}. Additionally, LLMs have been used as agents for hazard analysis for safety-critical systems due to their strong reasoning capabilities \cite{diemert2023largelanguagemodelsassist}. LLMs have also been used effectively in the construction industry for tasks such as reporting and generating construction progress reports \cite{AutoREPOLLM}. Other construction-related tasks such as compliance \cite{chen2024automated} and architectural detailing have also effectively benefited from LLMs  \cite{jang2024automated}. Leveraging previously known information to create informed ML models has shown significant promise in creating informed machine learning systems \cite{von2021informed}. By introducing prior knowledge explicitly in the model independent from the training data, ML algorithms have been shown to perform more effectively. Previous works have shown that NLP can be integrated in a similar fashion with unrelated learning algorithms to improve their performance\cite{kaplan2017beatingatarinaturallanguage}. The authors have previously explored semantic informed ML in the context of human activity recognition in the construction industry \cite{amani2024adaptiverobotperceptionconstruction}.
Given the above mentioned literature and proven potential, in this work we use LLMs for scaling and weighing the localization errors of the BIM families to strike a balance between object avoidance and path length. This is while the presented framework allows for transparent communication of the reason behind using certain scalings to enhance the pathfinding process, as well as having natural language communication ability to reason about mission planning.
\par

\section{Methodology}

This section outlines the building blocks of the proposed methodology, where APF and A* algorithms are fused to determine safe navigation paths. This is enabled by converting higher dimensional BIM models into graph-based, 2D floorplans, as described below.

\subsection{Graph Generation}
We propose a multi-heuristic A* (MHA*) algorithm to formulate a global path-finding process. Since A* is a graph traversal algorithm, the first step in the framework would be to generate a 2D grid representing the floor plan and walkable regions of the environment.

\begin{figure}[H]
    \centering
    \includegraphics[width=0.75\linewidth]{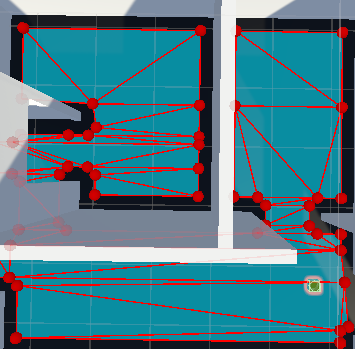}
    \caption{Graph Generation using Unity's Navigation Package}
    \label{NavMeshNode}
\end{figure}

 We use Python in conjunction with Unity Game Engine, which is a popular tool in the literature for robotics and BIM-based positioning studies due to its intuitive and visual 3D representations and ROS\# package \cite{de2019analysis}  \cite{stamford2014pathfinding}\cite{RSSprediciton}. It is important to note that the proposed framework does not require using a game engine as a necessary component, and it is leveraged here as a simulation and visualization tool only.
 Unity provides a navigation mesh function called NavMesh that saves the scene's geometry as convex polygons \cite{UnityDocs2023}, which has been tested previously in construction robotic path-planning studies \cite{halder2024robotic} with successful results. The function can detect walkable and un-walkable areas with high flexibility and uses the edges of the detected polygons to connect the traversable areas. An example of the NavMesh graph can be seen in Figure \ref{NavMeshNode}
Since the NavMesh function represents the geometry as convex polygons, typical graph traversal algorithms such as A* will calculate the shortest distance between the destination and the agent. This would result in the agent following along the edges of the polygons since the shortest distance would naturally cost less. 
\par
\begin{figure}[H]
    \centering
    \includegraphics[width=0.75\linewidth]{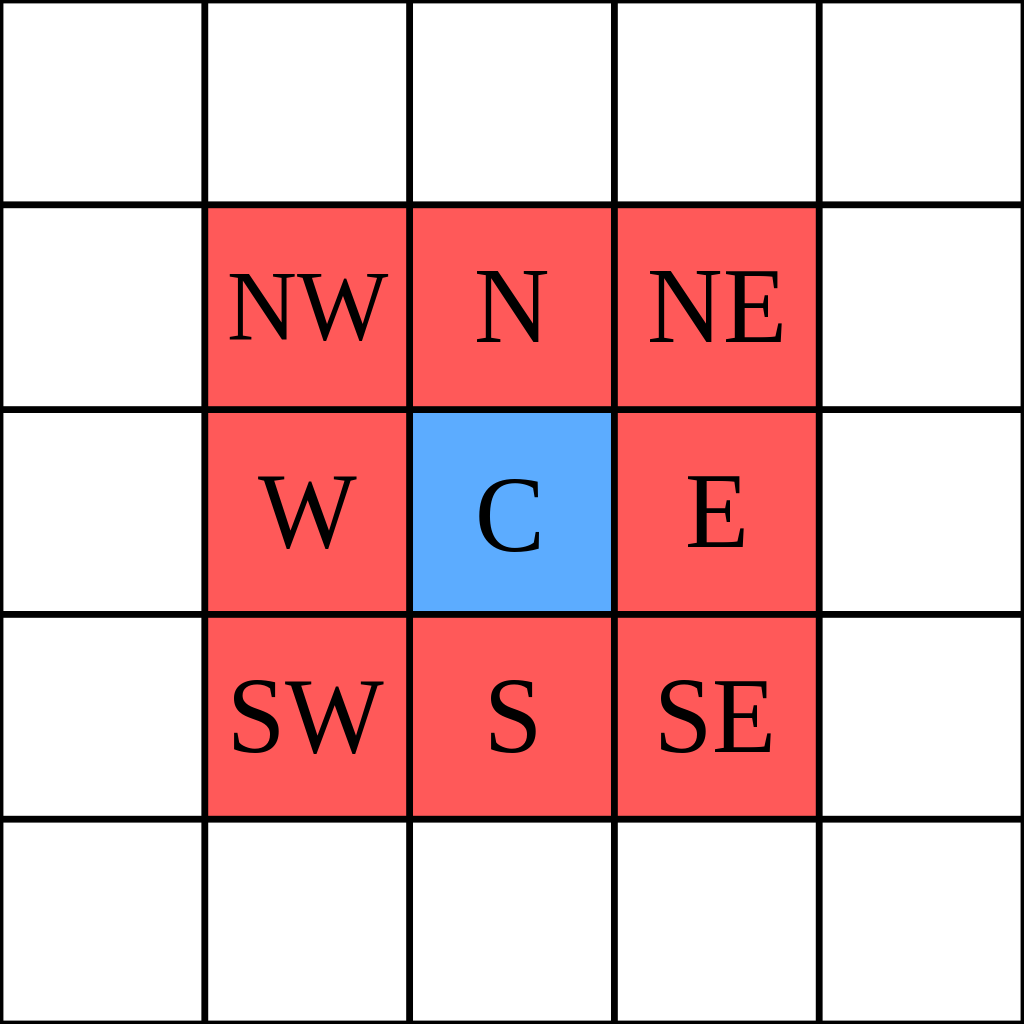}
    \caption{Moore's Neighborhood Graph }
    \label{Grid}
\end{figure}

To incorporate APFs for object-safe path planning, we need to be able to project the potential values across the environment. Even though having the NavMesh representation decreases computational costs while maintaining high functionality, we need to deviate from only using the edges of the polygons in our cost functions to incorporate APF into the system. We use the geometry of the floor to generate a "Moore's Neighborhood"-a grid similar to what is shown in Figure \ref{Grid}. 
\begin{figure}[H]
    \centering
    \includegraphics[width=0.75\linewidth]{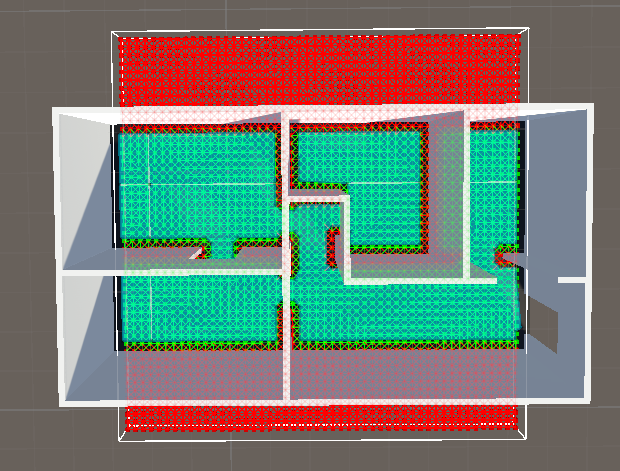}
    \caption{Generated Grid in Unity with the Moore's Neighborhood Node Connections}
    \label{Grid1}
\end{figure}
A Moore's Neighborhood grid, as depicted in Figure \ref{Grid}, is a 2-dimensional grid that grants 8 degrees of freedom to the agent. We chose this setup to avoid a "lawnmower" movement from the agent by connecting each node to all of its immediate neighbors \cite{zaitsev2017generalized}. 

Depending on the geometry and node sizes, this setup creates significantly more nodal connections than Unity's NavMesh (see Figure \ref{Grid1}). While this setup provides more granular and fine movements, it also increases the computational complexity of the mission. However, this is not a concern here since the frame rates and performance of the Unity client are not of significant importance in the proposed application compared to video-game settings. This situation increases the viability of developing alternative methodologies as opposed to the in-built NavMesh function for functional purposes. The determination of walkable and unwalkabale nodes are detected by the intersection of the grid objects with the walkable area of the NavMesh itself.

\subsection{Attractive Potential}
The attractive potential is calculated using the equation:
\begin{equation}
U_{\text{att}}(x, y) = \frac{1}{2} k_{\text{att}} \cdot D_{goal}
\end{equation}
where \( D_{goal} \) computes the distance between any grid point \((x, y)\) and the goal and \( k_{\text{att}} \) is the attractive coefficient. The farther we are from the goal, the stronger the potential field which facilitates more aggressive and direct path planning. Different scaling factors can be used to tune the function depending on applications and environments.
The distance can be calculated with different measurement approaches. In this paper, we examine the Euclidean distance:
\begin{equation}
    D_{Euclidian} = \sqrt{(x_{2}-x_{1})^2 + (y_{2}-y_{1})^2}
    \label{EQ2}
\end{equation}
where x and y are the coordinates of the start and end nodes.

\subsection{Repulsive Potential}
The repulsive potential around obstacles is given by an exponential decay function:
\begin{equation}
F_{\text{rep}}(x, y) =
k_{\text{rep}} \cdot e^{-D_{obstacle} }
\label{eq3}
\end{equation}
where D measures the distance to the nearest obstacle point, \( k_{\text{rep}} \) is the repulsive coefficient. Exponential decay functions have been used in previous works to create more smooth potentials \cite{peng2024smooth}. 
This will allow to create walkable regions in the floor plan generated from the BIM model.
\subsection{MHA* Search}
We utilize a MHA* graph traversal algorithm for our 
pathfinding problem. MHA* guarantees completeness and optimality even though the APF heuristic is potentially inadmissible. This is an advance from previous works that do not account for the admissibility of the APF. The proof for completeness and optimality can be further found in the original paper \cite{MHA*}. Our admissible heuristic is given by the Euclidean distance in Equation \ref{EQ2}. Our potentially inadmissible heuristic that is used for object avoidance is the summation of all the magnitude of the forces on each node minus the attractive force on the node:
\begin{equation}
    {F}_{\text{total}} = \sum_{i=1}^n k_{i} {F_{r}}_i - F_{a} 
\end{equation}
APF and Euclidean distance are now  used as our two heuristic functions for object-aware pathfinding. Each one of these heuristics will have its open set that will be chosen for state expansion in the search. 
\par
We also use a Gaussian blurring function that is convoluted with the potential field to get a smooth representation of the original potential field, which is a technique commonly used in 2D image processing \cite{haddad1991class}:
\begin{equation}
    G(x,y) = \frac{1}{2\pi\sigma^2} e^{-\frac{x^2 + y^2}{2\sigma^2}}
\end{equation}
\begin{figure}[h]
    \centering
    \includegraphics[width=\linewidth]{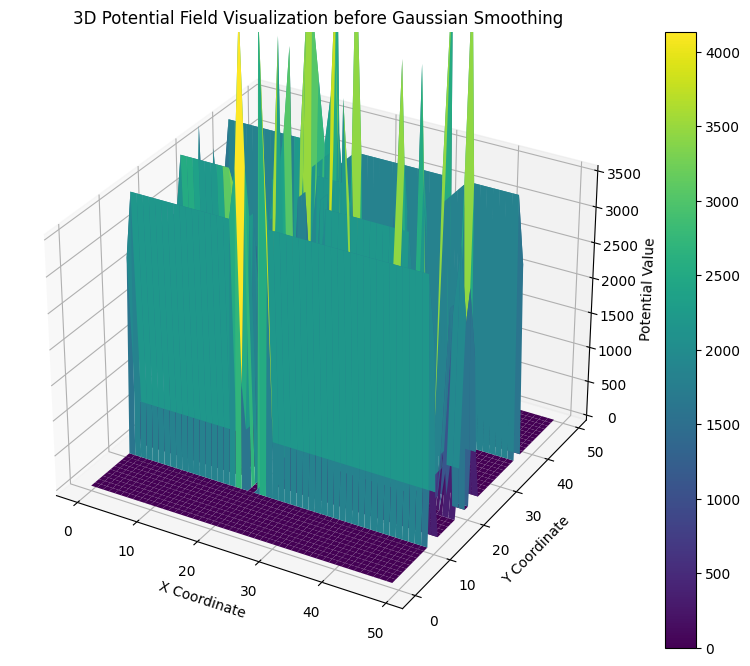}
    \includegraphics[width = \linewidth]{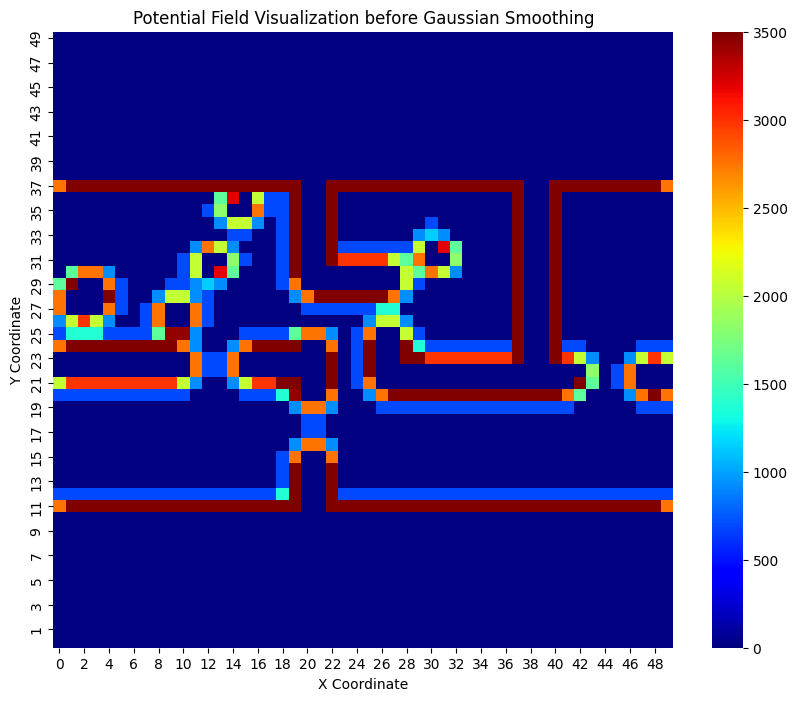}
    \caption{Potential field before Gaussian Smoothing}
    \label{BeforeSmoothing}
\end{figure}
\begin{figure}[h]
    \centering
    \includegraphics[width=\linewidth]{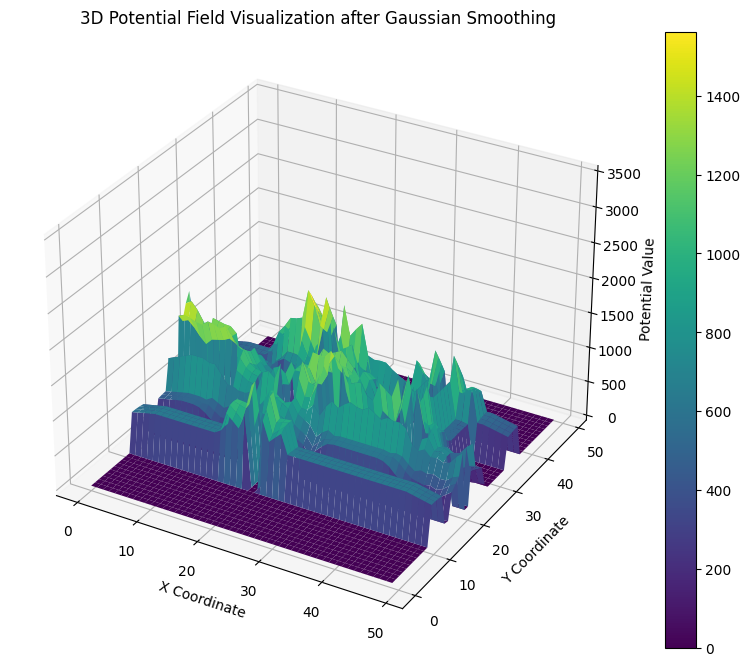}
    \includegraphics[width = \linewidth]{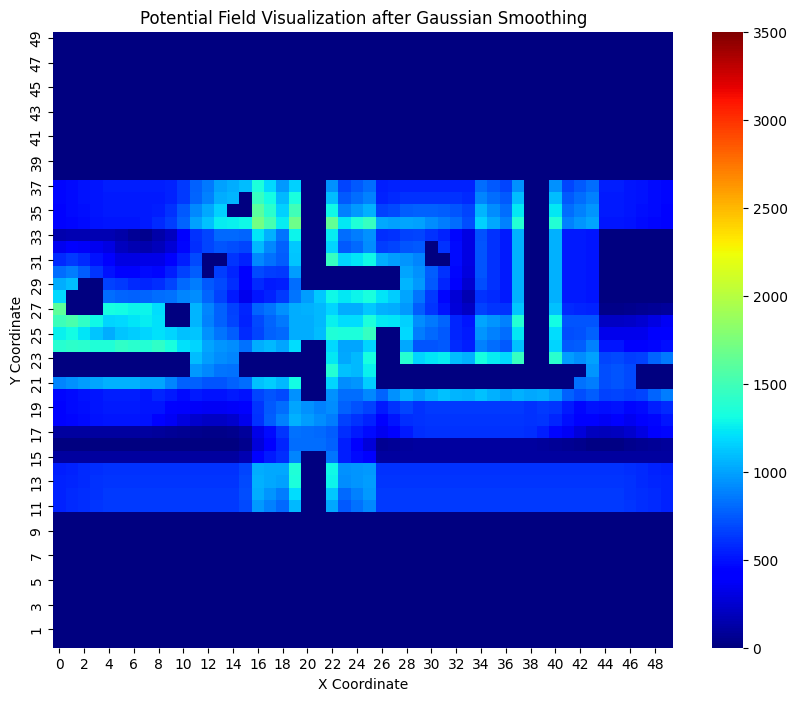}
    \caption{Potential Field after Gaussian Smoothing}
    \label{AfterSmoothing}
\end{figure}
Results of applying a convolution process over the Gaussian kernel can be seen in Figures \ref{BeforeSmoothing} and \ref{AfterSmoothing}. In these figures, the floorplan depicted in Figure \ref{Grid1} is transformed into a potential field emitted by the objects. A high potential value is represented in the red color with a maximum value of 3500, and the low potential values are represented with the blue color with the minimum of zero. The smoothing gives a more natural distribution of the potentials to further assist the algorithm in incorporating the potential better.
\par
Moreover, the Gaussian blur further stretches the effects of the potential field, acting as a second extension of the exponential function formulated in Equation \ref{eq3} to smooth out the initial function over the domain. 
The difference between not using and using the smoothing in pathfinding performance can be seen in Figures \ref{fig:WithoutSmoothing} and \ref{fig:WithSmoothing}.
\subsection{LLM}
Construction job processes have nuanced and esoteric nomenclature. Different companies and projects could use different terms for similar activities. This variation can cause errors and malfunctions in traditional NLP approaches. Previously, NLP was limited to regular expressions and keyword searching. With the rise of LLMs, NLP has the opportunity to use reasoning and context to better process prompts. In this setting, we introduce the integration of LLMs to be able to analyze BIM families names (i.e., a collection of elements with identical use, common parameters, and similar geometry) and other textual information. Given the dynamic nature of jobsites, certain objects can be moved or misplaced more frequently than others. Furthermore, certain localization errors can be communicated to the framework via the context prompt of the GPT. These flexibilities make an LLM integration an attractive extension to traditional construction robot mission planning. 
\par
\begin{figure}
    \centering
    \includegraphics[width=\linewidth]{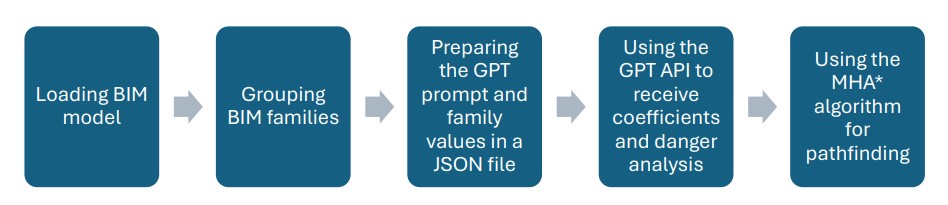}
    \caption{Framework of integrating LLM's into the A* Algorithm}
    \label{fig:GPTintegration}
\end{figure}

The framework script prompts the GPT using JSON files to supply the model with information. A list is created with every single family and object instance of the BIM file. The GPT is then prompted with instructions to identify danger levels given the description of each family. Then, these coefficients are applied to the A* algorithm. Different steps of the framework are shown in Figure \ref{fig:GPTintegration}.
\par
\section{Experiments and Validation}

We used a BIM model of an existing suite of rooms at SDSU (with its 2D plan appeared in the Figures 3-12)  to validate the pathfinding algorithm. Initially, an  (i.e., a geometry preserving file format used for interoperability) was exported from an RVT (i.e., Revit file) to generate a 3D representation of the space. To use it in Unity, we simply imported the FBX file and use the floor as the geometry for navigation. For the Python implementation, we generated a cross-section of the 3D file to generate a 2D map of the floor. In the following subsections, we delineate the differences between the output of MHA* (A* with APF) versus A* classic, and then expand on the BIM-LLM integration for the algorithm.
\subsubsection{MHA* vs classic A*}
\begin{figure}
    \centering
    \includegraphics[width=0.85\linewidth]{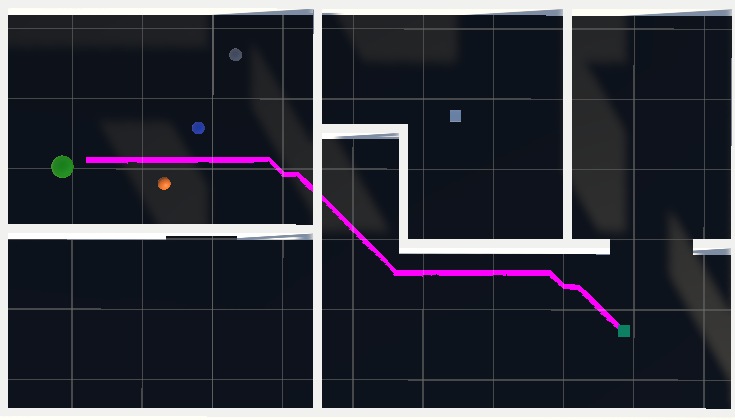}
    \caption{Naive A* Pathfinding. (Yellow: Chair, Blue: Robot, Gray: Grinder)}
    \label{NaiveA*}
\end{figure}

\begin{figure}[h]
    \centering
    \includegraphics[width=0.85\linewidth]{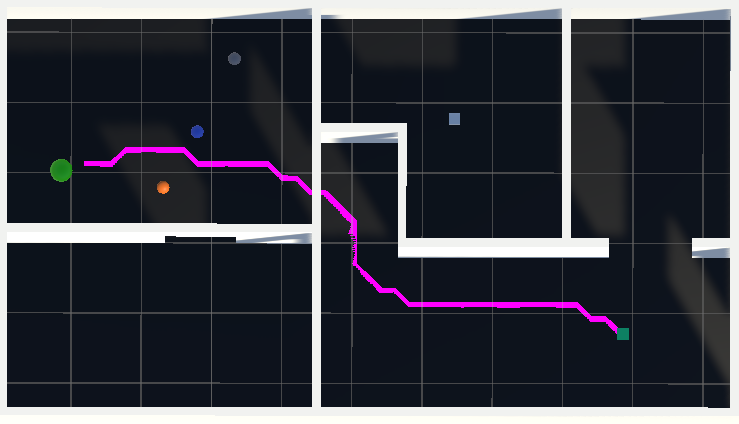}
    \caption{MHA* without Gaussian Smoothing}
    \label{fig:WithoutSmoothing}
\end{figure}

\begin{figure}[h]
    \centering
    \includegraphics[width=0.85\linewidth]{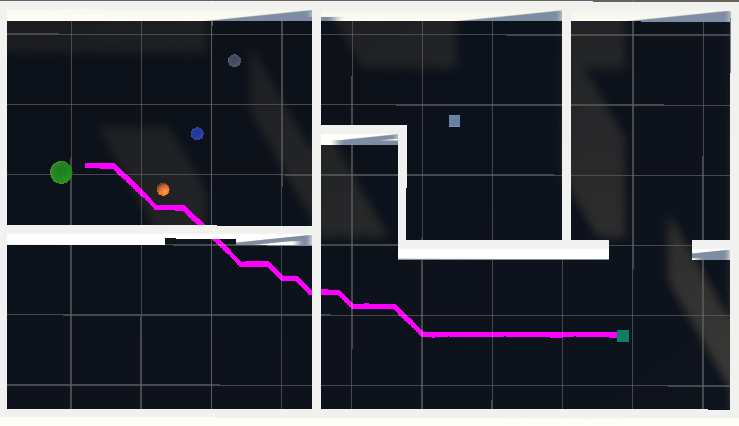}
    \caption{MHA* with Gaussian Smoothing}
    \label{fig:WithSmoothing}
\end{figure}

The output of any A* algorithm would be a list of node coordinates which are the immediate neighbors of the prior node. The metric that we used to evaluate the overall object avoidance of each algorithm is the average of the distance of each node from the closest un-walkable node (in this case the nodes are rendered un-walkable if they appear on an object) and the total count of the nodes needed to complete the path. We experimented with 1) a naive A* which only accounts for distance in its calculations, 2)  MHA* without Gaussian smoothing, and 3) MHA* approach with Gaussian smoothing. Figure \ref{NaiveA*} shows the naive A* path, Figure \ref{fig:WithoutSmoothing} shows the MHA* without Gaussian smoothing, and Figure \ref{fig:WithSmoothing} shows the path for the MHA* algorithm after Gaussian smoothing.

\begin{table}[h]
\centering
\begin{tabular}{|l|c|c|}
\hline
\textbf{Algorithm} & \textbf{Path Length} & \textbf{ADO} \\
\hline
Naive A* & 4.29 &  0.34 \\
\hline
MHA* without Gaussian Smoothing & 4.55 &  0.50 \\
\hline
MHA* with Gaussian Smoothing & 4.29 & 0.58\\
\hline
\end{tabular}
\caption{Comparison of Naive A* and A*-APF Fusion algorithms with path length and Average Distance to Obstacle (ADO)}
\label{tab:algorithm-comparison}
\end{table}

As it is shown in Table \ref{tab:algorithm-comparison}, with the MHA* method, the agent can reach the destination with the same path length of the naive algorithm. However, the method results in about 75\% increase in Average Distance to Obstacle (ADO) than the naive algorithm. This is a significant improvement in mission planning safety. Also, the effects of smoothing in reducing the path length while generating a smaller average distance to obstacles are noteworthy.
\par
In these scenarios, we gave the potential field a scaling factor of 0.5. The idea behind the scaling factor is to create a mechanism to be able to adjust the potential field generated from each object to affect the behavior of the path. However, this is not optimal since different objects can have similar perceived danger levels without considering their geometry, unpredictable movement, and cost. While the user can manually assign these values in the software, the textual data depending on the size of the BIM can become cumbersome, and 
near real-time changes to values depending on the environment and scenario can be unfeasible. Furthermore, given different nomenclatures regarding how to name BIM families depending on the organization, a direct scaling method is not effective. To address this, we propose to use LLMs to use the textual information embedded in the BIM model to dynamically weigh and scale the potential force emitted from the BIM families.

\subsubsection{LLM Integration}
To further integrate explainability and adaptability, we propose to use LLM as a reasoning agent to identify and detect danger levels. By using the names and descriptions of the families of the BIM model, the GPT can generate danger coefficients using prompts. The GPT can be used to generate either potential strength coefficients or cost scaling coefficients, or both in different implementations. In our framework, we use the GPT to generate scaling values between 0 to 1. 
\par
To integrate the LLM with the Unity game engine, we use the OpenAI GPT API. The model used was the GPT 3.5 turbo \cite{gpt35t}.
GPT 3.5- turbo was chosen due to its speed and efficacy. While other LLMs such as GPT-4 \cite{GPT4} or Gemini \cite{Gemini} could be used as well, they would require more resources and processing times due to their more complex architecture. Most modern GPTs such GPT 4o and o1 have orders of magnitude more parameters than the GPT 3.5-turbo. This increase in model size will increase cost and computational times, which can be problematic in practical applications. 
Depending on the number of families used in the BIM model, the scalar addition of the coefficients can exceed 1, which would cause the algorithm to fail. To counteract that, all the APF cost coefficients are scaled and normalized to a maximum of 0.5 for when all are added to each other.
\par
The framework also has an explanation section that communicates why each coefficient has been chosen to be able to better tune the model and have a transparent and safe operation. A formatted version of the GPT output in our test case is as follows:

\textbf{Object Evaluation and Danger Coefficients:}
\begin{itemize}
    \item \textbf{Basic Wall Interior Partition (x12):}
    \begin{itemize}
        \item Danger Level: 0.2
        \item Reasoning: Static objects, easily navigated around. Low danger level.
    \end{itemize}
    
    \item \textbf{Chair:}
    \begin{itemize}
        \item Danger Level: 0.5
        \item Reasoning: Potential obstruction, moderate risk.
    \end{itemize}
    
    \item \textbf{Other Robot:}
    \begin{itemize}
        \item Danger Level: 0.8
        \item Reasoning: Dynamic object with unpredictable movements. Higher risk.
    \end{itemize}
    
    \item \textbf{Grinder:}
    \begin{itemize}
        \item Danger Level: 0.9
        \item Reasoning: High-risk due to cutting capabilities and potential for serious harm. Highest danger level.
    \end{itemize}
\end{itemize}
\par
We can see transparent and valid reasoning for the danger levels of each object. After this text is returned, we use regular expressions (Regex) techniques \cite{regex} to parse the specific coefficients. These coefficients are then normalized and applied to the BIM families, preparing the potential fields for path planning. To evaluate this method we use two separate scenario initializations. 

\subsection{Scenario 1}
\begin{figure}[h]
    \centering

    \includegraphics[width=0.85\linewidth]{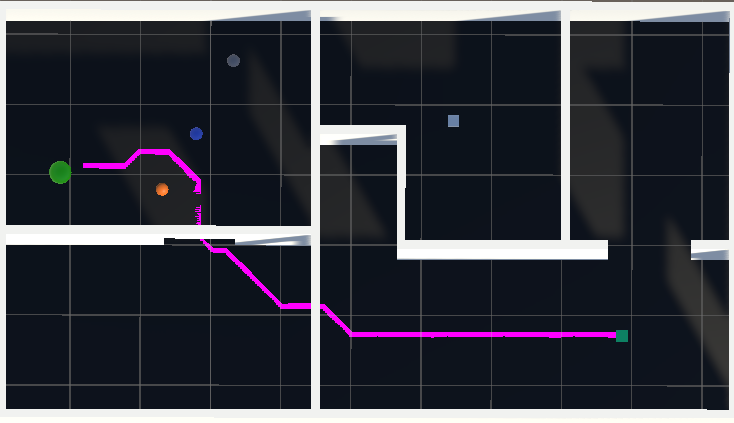}
    \caption{GPT-MHA* using GPT-3.5-turbo for Scaling Values Depending on the Semantic Evaluation of Danger Levels of the BIM Families}
    \label{fig:GPT-APF}
\end{figure}

We test the BIM layout from the previous experiments to compare LLM integration and the variants without sentiment analysis directly.
The results can be seen in Figure \ref{fig:GPT-APF} and Table \ref{tab:GPT-APF}. As it can be seen, the GPT weighing increases the average distance to approximately 0.63 meters which is about 80\% higher than the naive A* method. Figure \ref{fig:GPT-APF} and Table \ref{tab:GPT-APF} show the difference between the paths given the presence of the "chair" and the "walls" near the path and considering their relative danger level. 
\begin{table}[h]
\centering
\begin{tabular}{|l|c|c|}
\hline
\textbf{Algorithm} & \textbf{Path Length} & \textbf{ADO} \\
\hline
GPT-MHA* & 4.61 &  0.63\\
\hline
Naive A* & 4.29 &  0.34 \\
\hline
\end{tabular}
\caption{Results for the A*-APF Algorithm Integrated with GPT-3.5}
\label{tab:GPT-APF}
\end{table}
\subsection{Scenario 2}

\begin{figure}
    \centering
    \includegraphics[width=1\linewidth]{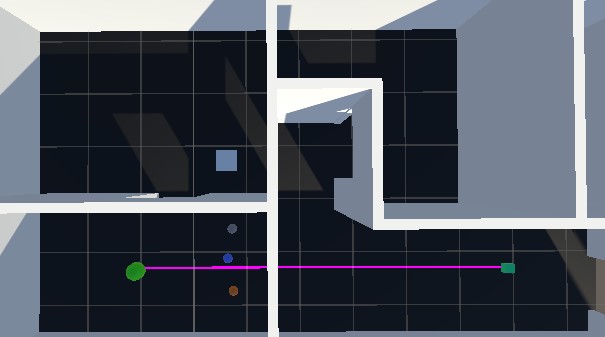}
    \caption{A* Path in Scenario 2}
    \label{fig:Scenario2}
\end{figure}

In this scenario, we place the objects close to the entrance of the nearest entry point to the room of interest. Given this setup, the safer route would be to enter the room above to enter through a side door as opposed to entering through the door with clustered obstacles.
Figure \ref{fig:Scenario2} gives the closest path given by A*. The naive A* will take the shortest path possible. However, the MHA* approach should avoid the cluster of nearby objects due to the repulsive heuristic. 

\begin{figure}[h]
    \centering
    \includegraphics[width=1\linewidth]{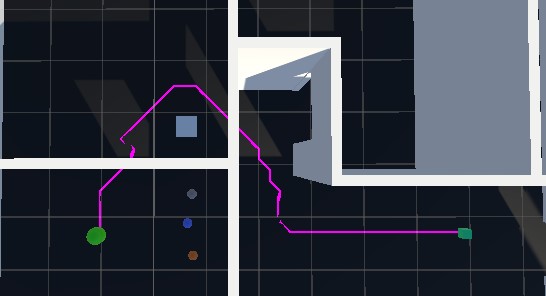}
    \caption{MHA* Path in Scenario 2}
    \label{fig:MHA*scenario2}
\end{figure}

Figure \ref{fig:MHA*scenario2} shows the path given by the proposed MHA* method. Table \ref{tab:Scenario2 stats} contains the statistics regarding the performance of the two different algorithms. We can see that the GPT-MHA* takes a longer path while maintaining a larger distance to obstacles on average as opposed to the naive A*. 

\begin{table}[h]
    \centering
    \begin{tabular}{|l|c|c|}
        \hline
        \textbf{Algorithm} & \textbf{Path Length} & \textbf{ADO} \\
        \hline
        GPT-MHA* & 5.35 & 0.28 \\
        \hline
        Naive A* & 3.4 & 0.2 \\
        \hline
    \end{tabular}
    \caption{Scenario 2 Path Statistics}
    \label{tab:Scenario2 stats}
\end{table}
\section{Discussion}
Path planning by leveraging digital twins has great potential for indoor environments. With the assumption that
near real-time representation of the work environment is reflected in the BIM world, this is even a more promising mechanism. As discussed earlier and shown in the literature, different environmental sensors such as RFID and camera sensors can prove to be object avoidance and mapping tools for the robot. However, with a better twinning of the locations of obstacles in a building environment, mission planning will be more likely to avoid collisions due to a more representative understanding of the environment. This advantage can further reduce the load from sensors used on the robot for object collision. This expenditure reduction can benefit other process performance, battery life, and implementation costs.
\par

The developed framework grants significant flexibility to users. Sensitive, high-value, or hazardous objects can each have their own specific potential strength and scaling factors. Certain rooms can be assigned with certain potential strengths and scaling factors to balance between the shortest distance and object avoidance even given the schedule and conditions of the environment. The balance between object avoidance and the shortest path needs to be determined by the use case of the robot's mission since higher weights for object avoidance can cause longer paths
\par
If the environment is unchanged, the A* algorithm executes once and a global path is acquired. This saves computational resources instead of calculating the path once and then using local path planning for better object avoidance. However, if the environment changes, the global path needs to be recalculated at every instance. Regardless, this is not detrimental to the robot since the computations are offloaded to a host computer. The simulation sends the robot move commands through ROS or any other API the users use. This protocol reduces the edge computing requirements of the robot and improves the latency of other functionalities that the robot needs to conduct.
\section{Conclusion and Future Work}
In conclusion, this work proposes a new path-planning technique that combines A* and APF using MHA* which is then further incorporated with an LLM for explainability and adaptability. We create a pipeline to use BIM spatial and semantic information to aid our path-planning logic with the aim of collision avoidance. While this method does not replace real-time object avoidance techniques, it creates an additional safety consideration in the planning phase for the robot before the mission starts. We can use the semantic and textual information from BIM using LLMs and incorporate them with classical AI algorithms. The GPT semantic analysis presented here can also provide transparency in the robot's decision-making, a crucial facet in the application of robots in the AEC industry.
\par
The proposed work focuses on implementing this navigation strategy in 2D for ground robots. This framework can be extended to 3D by incorporating the height axis into the heuristic calculation and grid generation. The grid will have a voxel representation with a repulsive and distance heuristic for each voxel. To ensure accurate potential calculation, accurate mesh representation of the objects is highly important. 


\par

For future work, we plan to create dynamic scalar adjustments depending on both semantic and environmental changes. We plan to incorporate 
online object avoidance techniques such as SLAM processes to create a more holistic framework. In this research, we did not explore the effects of the efficiency and convergence speed of the algorithm. Multi-resolution approaches have shown to be effective in reducing the computational needs of the A* algorithm by dynamically adjusting the search resolution of the algorithm. AMRA* is an anytime, multi-resolution, and multi-heuristic A* algorithm that can handle multiple heuristics and multiple resolutions which we plan to implement in the future \cite{Saxena_2022}.


%

\appendices

\ifCLASSOPTIONcaptionsoff
  \newpage
\fi

\bibliographystyle{plain}

\begin{thebibliography}{10}

\bibitem{VisualObjectAvoidance}
Wilbert~G. Aguilar, Verónica~P. Casaliglla, and José~L. Pólit.
\newblock Obstacle avoidance based-visual navigation for micro aerial vehicles.
\newblock {\em Electronics}, 6(1), 2017.

\bibitem{regex}
Alfred~V Aho.
\newblock Algorithms for finding patterns in strings, handbook of theoretical computer science (vol. a): algorithms and complexity, 1991.

\bibitem{MHA*}
Sandip Aine, Siddharth Swaminathan, Venkatraman Narayanan, Victor Hwang, and Maxim Likhachev.
\newblock Multi-heuristic a*.
\newblock {\em The International Journal of Robotics Research}, 35(1-3):224--243, 2016.

\bibitem{alayrac2022flamingovisuallanguagemodel}
Jean-Baptiste Alayrac, Jeff Donahue, Pauline Luc, Antoine Miech, Iain Barr, Yana Hasson, Karel Lenc, Arthur Mensch, Katie Millican, Malcolm Reynolds, Roman Ring, Eliza Rutherford, Serkan Cabi, Tengda Han, Zhitao Gong, Sina Samangooei, Marianne Monteiro, Jacob Menick, Sebastian Borgeaud, Andrew Brock, Aida Nematzadeh, Sahand Sharifzadeh, Mikolaj Binkowski, Ricardo Barreira, Oriol Vinyals, Andrew Zisserman, and Karen Simonyan.
\newblock Flamingo: a visual language model for few-shot learning, 2022.

\bibitem{amani2024adaptiverobotperceptionconstruction}
Mani Amani and Reza Akhavian.
\newblock Adaptive robot perception in construction environments using 4d bim, 2024.

\bibitem{ASCEBIMLOCALization}
Khashayar Asadi, Hariharan Ramshankar, Mojtaba Noghabaei, and Kevin Han.
\newblock Real-time image localization and registration with bim using perspective alignment for indoor monitoring of construction.
\newblock {\em Journal of Computing in Civil Engineering}, 33(5):04019031, 2019.

\bibitem{BagheriTrustExplain}
Elahe Bagheri, Joris De~Winter, and Bram Vanderborght .
\newblock Transparent interaction based learning for human-robot collaboration.
\newblock {\em Frontiers in Robotics and AI}, 9, 2022.

\bibitem{TwinPath}
Qiangwei Bao, Pai Zheng, and Sheng Dai.
\newblock A digital twin-driven dynamic path planning approach for multiple automatic guided vehicles based on deep reinforcement learning.
\newblock {\em Proceedings of the Institution of Mechanical Engineers, Part B: Journal of Engineering Manufacture}, 238(4):488--499, 2024.

\bibitem{chen2014integration}
Hung-Ming Chen and Ting-Yu Chang.
\newblock Integration of augmented reality and indoor positioning technologies for on-site viewing of bim information.
\newblock In {\em ISARC. Proceedings of the International Symposium on Automation and Robotics in Construction}, volume~31, page~1. IAARC Publications, 2014.

\bibitem{chen2024automated}
Nanjiang Chen, Xuhui Lin, Hai Jiang, and Yi~An.
\newblock Automated building information modeling compliance check through a large language model combined with deep learning and ontology.
\newblock {\em Buildings}, 14(7):1983, 2024.

\bibitem{costin2015fusing}
Aaron~M Costin and Jochen Teizer.
\newblock Fusing passive rfid and bim for increased accuracy in indoor localization.
\newblock {\em Visualization in Engineering}, 3:1--20, 2015.

\bibitem{creswell2022faithfulreasoningusinglarge}
Antonia Creswell and Murray Shanahan.
\newblock Faithful reasoning using large language models, 2022.

\bibitem{NikolayLLMA}
Zhirui Dai, Arash Asgharivaskasi, Thai Duong, Shusen Lin, Maria-Elizabeth Tzes, George Pappas, and Nikolay Atanasov.
\newblock Optimal scene graph planning with large language model guidance.
\newblock In {\em 2024 IEEE International Conference on Robotics and Automation (ICRA)}, pages 14062--14069, 2024.

\bibitem{de2019analysis}
Mirella Santos~Pessoa De~Melo, Jos{\'e}~Gomes da~Silva~Neto, Pedro Jorge~Lima Da~Silva, Jo{\~a}o Marcelo Xavier~Natario Teixeira, and Veronica Teichrieb.
\newblock Analysis and comparison of robotics 3d simulators.
\newblock In {\em 2019 21st Symposium on Virtual and Augmented Reality (SVR)}, pages 242--251. IEEE, 2019.

\bibitem{deng2017bim}
Y~Deng, H~Hong, H~Deng, and H~Luo.
\newblock Bim-based indoor positioning technology using a monocular camera, isarc.
\newblock In {\em Proceedings of the International Symposium on Automation and Robotics in Construction, Vilnius Gediminas Technical University, Department of Construction Economics \& Property}, 2017.

\bibitem{diemert2023largelanguagemodelsassist}
Simon Diemert and Jens~H Weber.
\newblock Can large language models assist in hazard analysis?, 2023.

\bibitem{du2020multi}
Wei Du, Fahad Islam, and Maxim Likhachev.
\newblock Multi-resolution a.
\newblock In {\em Proceedings of the International Symposium on Combinatorial Search}, volume~11, pages 29--37, 2020.

\bibitem{WeightedAstar}
Rüdiger Ebendt and Rolf Drechsler.
\newblock Weighted a∗ search – unifying view and application.
\newblock {\em Artificial Intelligence}, 173(14):1310--1342, 2009.

\bibitem{EMAMINEJAD2022104298}
Newsha Emaminejad and Reza Akhavian.
\newblock Trustworthy ai and robotics: Implications for the aec industry.
\newblock {\em Automation in Construction}, 139:104298, 2022.

\bibitem{TrustNewsha}
Newsha Emaminejad, Lisa Kath, and Reza Akhavian.
\newblock Assessing trust in construction ai-powered collaborative robots using structural equation modeling.
\newblock {\em Journal of Computing in Civil Engineering}, 38(3):04024011, 2024.

\bibitem{haddad1991class}
Richard~A Haddad, Ali~N Akansu, et~al.
\newblock A class of fast gaussian binomial filters for speech and image processing.
\newblock {\em IEEE Transactions on Signal Processing}, 39(3):723--727, 1991.

\bibitem{halder2024robotic}
Srijeet Halder, Kereshmeh Afsari, and Abiola Akanmu.
\newblock A robotic cyber-physical system for automated reality capture and visualization in construction progress monitoring, 2024.

\bibitem{RSSprediciton}
Hamid Hosseini, Mohammad Taleai, and Sisi Zlatanova.
\newblock Nsga-ii based optimal wi-fi access point placement for indoor positioning: A bim-based rss prediction.
\newblock {\em Automation in Construction}, 152:104897, 2023.

\bibitem{jang2024automated}
Suhyung Jang, Ghang Lee, Jiseok Oh, Junghun Lee, and Bonsang Koo.
\newblock Automated detailing of exterior walls using nadia: Natural-language-based architectural detailing through interaction with ai.
\newblock {\em Advanced Engineering Informatics}, 61:102532, 2024.

\bibitem{kaplan2017beatingatarinaturallanguage}
Russell Kaplan, Christopher Sauer, and Alexander Sosa.
\newblock Beating atari with natural language guided reinforcement learning, 2017.

\bibitem{Anytime}
Maxim Likhachev, Geoffrey~J Gordon, and Sebastian Thrun.
\newblock Ara* : Anytime a* with provable bounds on sub-optimality.
\newblock In S.~Thrun, L.~Saul, and B.~Sch\"{o}lkopf, editors, {\em Advances in Neural Information Processing Systems}, volume~16. MIT Press, 2003.

\bibitem{liu2017scene}
Mengyun Liu, Ruizhi Chen, Deren Li, Yujin Chen, Guangyi Guo, Zhipeng Cao, and Yuanjin Pan.
\newblock Scene recognition for indoor localization using a multi-sensor fusion approach.
\newblock {\em Sensors}, 17(12):2847, 2017.

\bibitem{RefA2}
Zhihai Liu, Hanbin Liu, Zhenguo Lu, and Qingliang Zeng.
\newblock A dynamic fusion pathfinding algorithm using delaunay triangulation and improved a-star for mobile robots.
\newblock {\em IEEE Access}, 9:20602--20621, 2021.

\bibitem{RefA1}
Min Luo, Xiaorong Hou, and Jing Yang.
\newblock Surface optimal path planning using an extended dijkstra algorithm.
\newblock {\em IEEE Access}, 8:147827--147838, 2020.

\bibitem{moura2021bim}
Mateus~Sanches Moura, Carlos Rizzo, and Daniel Serrano.
\newblock Bim-based localization and mapping for mobile robots in construction.
\newblock In {\em 2021 IEEE international conference on autonomous robot systems and competitions (ICARSC)}, pages 12--18. IEEE, 2021.

\bibitem{GPT4}
OpenAI, Josh Achiam, Steven Adler, Sandhini Agarwal, Lama Ahmad, Ilge Akkaya, Florencia~Leoni Aleman, Diogo Almeida, Janko Altenschmidt, Sam Altman, Shyamal Anadkat, Red Avila, Igor Babuschkin, Suchir Balaji, Valerie Balcom, Paul Baltescu, Haiming Bao, Mohammad Bavarian, Jeff Belgum, Irwan Bello, Jake Berdine, Gabriel Bernadett-Shapiro, Christopher Berner, Lenny Bogdonoff, Oleg Boiko, Madelaine Boyd, Anna-Luisa Brakman, Greg Brockman, Tim Brooks, Miles Brundage, Kevin Button, Trevor Cai, Rosie Campbell, Andrew Cann, Brittany Carey, Chelsea Carlson, Rory Carmichael, Brooke Chan, Che Chang, Fotis Chantzis, Derek Chen, Sully Chen, Ruby Chen, Jason Chen, Mark Chen, Ben Chess, Chester Cho, Casey Chu, Hyung~Won Chung, Dave Cummings, Jeremiah Currier, Yunxing Dai, Cory Decareaux, Thomas Degry, Noah Deutsch, Damien Deville, Arka Dhar, David Dohan, Steve Dowling, Sheila Dunning, Adrien Ecoffet, Atty Eleti, Tyna Eloundou, David Farhi, Liam Fedus, Niko Felix, Simón~Posada Fishman, Juston Forte, Isabella Fulford, Leo
  Gao, Elie Georges, Christian Gibson, Vik Goel, Tarun Gogineni, Gabriel Goh, Rapha Gontijo-Lopes, Jonathan Gordon, Morgan Grafstein, Scott Gray, Ryan Greene, Joshua Gross, Shixiang~Shane Gu, Yufei Guo, Chris Hallacy, Jesse Han, Jeff Harris, Yuchen He, Mike Heaton, Johannes Heidecke, Chris Hesse, Alan Hickey, Wade Hickey, Peter Hoeschele, Brandon Houghton, Kenny Hsu, Shengli Hu, Xin Hu, Joost Huizinga, Shantanu Jain, Shawn Jain, Joanne Jang, Angela Jiang, Roger Jiang, Haozhun Jin, Denny Jin, Shino Jomoto, Billie Jonn, Heewoo Jun, Tomer Kaftan, Łukasz Kaiser, Ali Kamali, Ingmar Kanitscheider, Nitish~Shirish Keskar, Tabarak Khan, Logan Kilpatrick, Jong~Wook Kim, Christina Kim, Yongjik Kim, Jan~Hendrik Kirchner, Jamie Kiros, Matt Knight, Daniel Kokotajlo, Łukasz Kondraciuk, Andrew Kondrich, Aris Konstantinidis, Kyle Kosic, Gretchen Krueger, Vishal Kuo, Michael Lampe, Ikai Lan, Teddy Lee, Jan Leike, Jade Leung, Daniel Levy, Chak~Ming Li, Rachel Lim, Molly Lin, Stephanie Lin, Mateusz Litwin, Theresa Lopez, Ryan
  Lowe, Patricia Lue, Anna Makanju, Kim Malfacini, Sam Manning, Todor Markov, Yaniv Markovski, Bianca Martin, Katie Mayer, Andrew Mayne, Bob McGrew, Scott~Mayer McKinney, Christine McLeavey, Paul McMillan, Jake McNeil, David Medina, Aalok Mehta, Jacob Menick, Luke Metz, Andrey Mishchenko, Pamela Mishkin, Vinnie Monaco, Evan Morikawa, Daniel Mossing, Tong Mu, Mira Murati, Oleg Murk, David Mély, Ashvin Nair, Reiichiro Nakano, Rajeev Nayak, Arvind Neelakantan, Richard Ngo, Hyeonwoo Noh, Long Ouyang, Cullen O'Keefe, Jakub Pachocki, Alex Paino, Joe Palermo, Ashley Pantuliano, Giambattista Parascandolo, Joel Parish, Emy Parparita, Alex Passos, Mikhail Pavlov, Andrew Peng, Adam Perelman, Filipe de~Avila Belbute~Peres, Michael Petrov, Henrique~Ponde de~Oliveira~Pinto, Michael, Pokorny, Michelle Pokrass, Vitchyr~H. Pong, Tolly Powell, Alethea Power, Boris Power, Elizabeth Proehl, Raul Puri, Alec Radford, Jack Rae, Aditya Ramesh, Cameron Raymond, Francis Real, Kendra Rimbach, Carl Ross, Bob Rotsted, Henri Roussez,
  Nick Ryder, Mario Saltarelli, Ted Sanders, Shibani Santurkar, Girish Sastry, Heather Schmidt, David Schnurr, John Schulman, Daniel Selsam, Kyla Sheppard, Toki Sherbakov, Jessica Shieh, Sarah Shoker, Pranav Shyam, Szymon Sidor, Eric Sigler, Maddie Simens, Jordan Sitkin, Katarina Slama, Ian Sohl, Benjamin Sokolowsky, Yang Song, Natalie Staudacher, Felipe~Petroski Such, Natalie Summers, Ilya Sutskever, Jie Tang, Nikolas Tezak, Madeleine~B. Thompson, Phil Tillet, Amin Tootoonchian, Elizabeth Tseng, Preston Tuggle, Nick Turley, Jerry Tworek, Juan Felipe~Cerón Uribe, Andrea Vallone, Arun Vijayvergiya, Chelsea Voss, Carroll Wainwright, Justin~Jay Wang, Alvin Wang, Ben Wang, Jonathan Ward, Jason Wei, CJ~Weinmann, Akila Welihinda, Peter Welinder, Jiayi Weng, Lilian Weng, Matt Wiethoff, Dave Willner, Clemens Winter, Samuel Wolrich, Hannah Wong, Lauren Workman, Sherwin Wu, Jeff Wu, Michael Wu, Kai Xiao, Tao Xu, Sarah Yoo, Kevin Yu, Qiming Yuan, Wojciech Zaremba, Rowan Zellers, Chong Zhang, Marvin Zhang, Shengjia
  Zhao, Tianhao Zheng, Juntang Zhuang, William Zhuk, and Barret Zoph.
\newblock Gpt-4 technical report, 2024.

\bibitem{peng2024smooth}
Bo~Peng, Lingke Zhang, and Rong Xiong.
\newblock Smooth path planning with subharmonic artificial potential field.
\newblock {\em arXiv preprint arXiv:2402.11601}, 2024.

\bibitem{AutoREPOLLM}
Hongxu Pu, Xincong Yang, Jing Li, and Runhao Guo.
\newblock Autorepo: A general framework for multimodal llm-based automated construction reporting.
\newblock {\em Expert Systems with Applications}, 255:124601, 2024.

\bibitem{MDPIdronesPath}
Hongwei Qin, Shiliang Shao, Ting Wang, Xiaotian Yu, Yi~Jiang, and Zonghan Cao.
\newblock Review of autonomous path planning algorithms for mobile robots.
\newblock {\em Drones}, 7(3), 2023.

\bibitem{russell2016artificial}
Stuart~J Russell and Peter Norvig.
\newblock {\em Artificial intelligence: a modern approach}.
\newblock Pearson, 2016.

\bibitem{sang2021hybrid}
Hongqiang Sang, Yusong You, Xiujun Sun, Ying Zhou, and Fen Liu.
\newblock The hybrid path planning algorithm based on improved a* and artificial potential field for unmanned surface vehicle formations.
\newblock {\em Ocean Engineering}, 223:108709, 2021.

\bibitem{Saxena_2022}
Dhruv~Mauria Saxena, Tushar Kusnur, and Maxim Likhachev.
\newblock Amra*: Anytime multi-resolution multi-heuristic a*.
\newblock In {\em 2022 International Conference on Robotics and Automation (ICRA)}. IEEE, May 2022.

\bibitem{schmidt2014automatic}
Bernard Schmidt and Lihui Wang.
\newblock Automatic work objects calibration via a global--local camera system.
\newblock {\em Robotics and Computer-Integrated Manufacturing}, 30(6):678--683, 2014.

\bibitem{Explainability}
Rossitza Setchi, Maryam~Banitalebi Dehkordi, and Juwairiya~Siraj Khan.
\newblock Explainable robotics in human-robot interactions.
\newblock {\em Procedia Computer Science}, 176:3057--3066, 2020.
\newblock Knowledge-Based and Intelligent Information \& Engineering Systems: Proceedings of the 24th International Conference KES2020.

\bibitem{stamford2014pathfinding}
John Stamford, Arjab~Singh Khuman, Jenny Carter, and Samad Ahmadi.
\newblock Pathfinding in partially explored games environments: The application of the a* algorithm with occupancy grids in unity3d.
\newblock In {\em 2014 14th UK Workshop on Computational Intelligence (UKCI)}, pages 1--6. IEEE, 2014.

\bibitem{Gemini}
Gemini Team, Rohan Anil, Sebastian Borgeaud, Jean-Baptiste Alayrac, Jiahui Yu, Radu Soricut, Johan Schalkwyk, Andrew~M. Dai, Anja Hauth, Katie Millican, David Silver, Melvin Johnson, Ioannis Antonoglou, Julian Schrittwieser, Amelia Glaese, Jilin Chen, Emily Pitler, Timothy Lillicrap, Angeliki Lazaridou, Orhan Firat, James Molloy, Michael Isard, Paul~R. Barham, Tom Hennigan, Benjamin Lee, Fabio Viola, Malcolm Reynolds, Yuanzhong Xu, Ryan Doherty, Eli Collins, Clemens Meyer, Eliza Rutherford, Erica Moreira, Kareem Ayoub, Megha Goel, Jack Krawczyk, Cosmo Du, Ed~Chi, Heng-Tze Cheng, Eric Ni, Purvi Shah, Patrick Kane, Betty Chan, Manaal Faruqui, Aliaksei Severyn, Hanzhao Lin, YaGuang Li, Yong Cheng, Abe Ittycheriah, Mahdis Mahdieh, Mia Chen, Pei Sun, Dustin Tran, Sumit Bagri, Balaji Lakshminarayanan, Jeremiah Liu, Andras Orban, Fabian Güra, Hao Zhou, Xinying Song, Aurelien Boffy, Harish Ganapathy, Steven Zheng, HyunJeong Choe, Ágoston Weisz, Tao Zhu, Yifeng Lu, Siddharth Gopal, Jarrod Kahn, Maciej Kula, Jeff
  Pitman, Rushin Shah, Emanuel Taropa, Majd~Al Merey, Martin Baeuml, Zhifeng Chen, Laurent~El Shafey, Yujing Zhang, Olcan Sercinoglu, George Tucker, Enrique Piqueras, Maxim Krikun, Iain Barr, Nikolay Savinov, Ivo Danihelka, Becca Roelofs, Anaïs White, Anders Andreassen, Tamara von Glehn, Lakshman Yagati, Mehran Kazemi, Lucas Gonzalez, Misha Khalman, Jakub Sygnowski, Alexandre Frechette, Charlotte Smith, Laura Culp, Lev Proleev, Yi~Luan, Xi~Chen, James Lottes, Nathan Schucher, Federico Lebron, Alban Rrustemi, Natalie Clay, Phil Crone, Tomas Kocisky, Jeffrey Zhao, Bartek Perz, Dian Yu, Heidi Howard, Adam Bloniarz, Jack~W. Rae, Han Lu, Laurent Sifre, Marcello Maggioni, Fred Alcober, Dan Garrette, Megan Barnes, Shantanu Thakoor, Jacob Austin, Gabriel Barth-Maron, William Wong, Rishabh Joshi, Rahma Chaabouni, Deeni Fatiha, Arun Ahuja, Gaurav~Singh Tomar, Evan Senter, Martin Chadwick, Ilya Kornakov, Nithya Attaluri, Iñaki Iturrate, Ruibo Liu, Yunxuan Li, Sarah Cogan, Jeremy Chen, Chao Jia, Chenjie Gu, Qiao Zhang,
  Jordan Grimstad, Ale~Jakse Hartman, Xavier Garcia, Thanumalayan~Sankaranarayana Pillai, Jacob Devlin, Michael Laskin, Diego de~Las~Casas, Dasha Valter, Connie Tao, Lorenzo Blanco, Adrià~Puigdomènech Badia, David Reitter, Mianna Chen, Jenny Brennan, Clara Rivera, Sergey Brin, Shariq Iqbal, Gabriela Surita, Jane Labanowski, Abhi Rao, Stephanie Winkler, Emilio Parisotto, Yiming Gu, Kate Olszewska, Ravi Addanki, Antoine Miech, Annie Louis, Denis Teplyashin, Geoff Brown, Elliot Catt, Jan Balaguer, Jackie Xiang, Pidong Wang, Zoe Ashwood, Anton Briukhov, Albert Webson, Sanjay Ganapathy, Smit Sanghavi, Ajay Kannan, Ming-Wei Chang, Axel Stjerngren, Josip Djolonga, Yuting Sun, Ankur Bapna, Matthew Aitchison, Pedram Pejman, Henryk Michalewski, Tianhe Yu, Cindy Wang, Juliette Love, Junwhan Ahn, Dawn Bloxwich, Kehang Han, Peter Humphreys, Thibault Sellam, James Bradbury, Varun Godbole, Sina Samangooei, Bogdan Damoc, Alex Kaskasoli, Sébastien M.~R. Arnold, Vijay Vasudevan, Shubham Agrawal, Jason Riesa, Dmitry
  Lepikhin, Richard Tanburn, Srivatsan Srinivasan, Hyeontaek Lim, Sarah Hodkinson, Pranav Shyam, Johan Ferret, Steven Hand, Ankush Garg, Tom~Le Paine, Jian Li, Yujia Li, Minh Giang, Alexander Neitz, Zaheer Abbas, Sarah York, Machel Reid, Elizabeth Cole, Aakanksha Chowdhery, Dipanjan Das, Dominika Rogozińska, Vitaliy Nikolaev, Pablo Sprechmann, Zachary Nado, Lukas Zilka, Flavien Prost, Luheng He, Marianne Monteiro, Gaurav Mishra, Chris Welty, Josh Newlan, Dawei Jia, Miltiadis Allamanis, Clara~Huiyi Hu, Raoul de~Liedekerke, Justin Gilmer, Carl Saroufim, Shruti Rijhwani, Shaobo Hou, Disha Shrivastava, Anirudh Baddepudi, Alex Goldin, Adnan Ozturel, Albin Cassirer, Yunhan Xu, Daniel Sohn, Devendra Sachan, Reinald~Kim Amplayo, Craig Swanson, Dessie Petrova, Shashi Narayan, Arthur Guez, Siddhartha Brahma, Jessica Landon, Miteyan Patel, Ruizhe Zhao, Kevin Villela, Luyu Wang, Wenhao Jia, Matthew Rahtz, Mai Giménez, Legg Yeung, James Keeling, Petko Georgiev, Diana Mincu, Boxi Wu, Salem Haykal, Rachel Saputro, Kiran
  Vodrahalli, James Qin, Zeynep Cankara, Abhanshu Sharma, Nick Fernando, Will Hawkins, Behnam Neyshabur, Solomon Kim, Adrian Hutter, Priyanka Agrawal, Alex Castro-Ros, George van~den Driessche, Tao Wang, Fan Yang, Shuo yiin Chang, Paul Komarek, Ross McIlroy, Mario Lučić, Guodong Zhang, Wael Farhan, Michael Sharman, Paul Natsev, Paul Michel, Yamini Bansal, Siyuan Qiao, Kris Cao, Siamak Shakeri, Christina Butterfield, Justin Chung, Paul~Kishan Rubenstein, Shivani Agrawal, Arthur Mensch, Kedar Soparkar, Karel Lenc, Timothy Chung, Aedan Pope, Loren Maggiore, Jackie Kay, Priya Jhakra, Shibo Wang, Joshua Maynez, Mary Phuong, Taylor Tobin, Andrea Tacchetti, Maja Trebacz, Kevin Robinson, Yash Katariya, Sebastian Riedel, Paige Bailey, Kefan Xiao, Nimesh Ghelani, Lora Aroyo, Ambrose Slone, Neil Houlsby, Xuehan Xiong, Zhen Yang, Elena Gribovskaya, Jonas Adler, Mateo Wirth, Lisa Lee, Music Li, Thais Kagohara, Jay Pavagadhi, Sophie Bridgers, Anna Bortsova, Sanjay Ghemawat, Zafarali Ahmed, Tianqi Liu, Richard Powell,
  Vijay Bolina, Mariko Iinuma, Polina Zablotskaia, James Besley, Da-Woon Chung, Timothy Dozat, Ramona Comanescu, Xiance Si, Jeremy Greer, Guolong Su, Martin Polacek, Raphaël~Lopez Kaufman, Simon Tokumine, Hexiang Hu, Elena Buchatskaya, Yingjie Miao, Mohamed Elhawaty, Aditya Siddhant, Nenad Tomasev, Jinwei Xing, Christina Greer, Helen Miller, Shereen Ashraf, Aurko Roy, Zizhao Zhang, Ada Ma, Angelos Filos, Milos Besta, Rory Blevins, Ted Klimenko, Chih-Kuan Yeh, Soravit Changpinyo, Jiaqi Mu, Oscar Chang, Mantas Pajarskas, Carrie Muir, Vered Cohen, Charline~Le Lan, Krishna Haridasan, Amit Marathe, Steven Hansen, Sholto Douglas, Rajkumar Samuel, Mingqiu Wang, Sophia Austin, Chang Lan, Jiepu Jiang, Justin Chiu, Jaime~Alonso Lorenzo, Lars~Lowe Sjösund, Sébastien Cevey, Zach Gleicher, Thi Avrahami, Anudhyan Boral, Hansa Srinivasan, Vittorio Selo, Rhys May, Konstantinos Aisopos, Léonard Hussenot, Livio~Baldini Soares, Kate Baumli, Michael~B. Chang, Adrià Recasens, Ben Caine, Alexander Pritzel, Filip Pavetic,
  Fabio Pardo, Anita Gergely, Justin Frye, Vinay Ramasesh, Dan Horgan, Kartikeya Badola, Nora Kassner, Subhrajit Roy, Ethan Dyer, Víctor~Campos Campos, Alex Tomala, Yunhao Tang, Dalia~El Badawy, Elspeth White, Basil Mustafa, Oran Lang, Abhishek Jindal, Sharad Vikram, Zhitao Gong, Sergi Caelles, Ross Hemsley, Gregory Thornton, Fangxiaoyu Feng, Wojciech Stokowiec, Ce~Zheng, Phoebe Thacker, Çağlar Ünlü, Zhishuai Zhang, Mohammad Saleh, James Svensson, Max Bileschi, Piyush Patil, Ankesh Anand, Roman Ring, Katerina Tsihlas, Arpi Vezer, Marco Selvi, Toby Shevlane, Mikel Rodriguez, Tom Kwiatkowski, Samira Daruki, Keran Rong, Allan Dafoe, Nicholas FitzGerald, Keren Gu-Lemberg, Mina Khan, Lisa~Anne Hendricks, Marie Pellat, Vladimir Feinberg, James Cobon-Kerr, Tara Sainath, Maribeth Rauh, Sayed~Hadi Hashemi, Richard Ives, Yana Hasson, Eric Noland, Yuan Cao, Nathan Byrd, Le~Hou, Qingze Wang, Thibault Sottiaux, Michela Paganini, Jean-Baptiste Lespiau, Alexandre Moufarek, Samer Hassan, Kaushik Shivakumar, Joost van
  Amersfoort, Amol Mandhane, Pratik Joshi, Anirudh Goyal, Matthew Tung, Andrew Brock, Hannah Sheahan, Vedant Misra, Cheng Li, Nemanja Rakićević, Mostafa Dehghani, Fangyu Liu, Sid Mittal, Junhyuk Oh, Seb Noury, Eren Sezener, Fantine Huot, Matthew Lamm, Nicola~De Cao, Charlie Chen, Sidharth Mudgal, Romina Stella, Kevin Brooks, Gautam Vasudevan, Chenxi Liu, Mainak Chain, Nivedita Melinkeri, Aaron Cohen, Venus Wang, Kristie Seymore, Sergey Zubkov, Rahul Goel, Summer Yue, Sai Krishnakumaran, Brian Albert, Nate Hurley, Motoki Sano, Anhad Mohananey, Jonah Joughin, Egor Filonov, Tomasz Kępa, Yomna Eldawy, Jiawern Lim, Rahul Rishi, Shirin Badiezadegan, Taylor Bos, Jerry Chang, Sanil Jain, Sri Gayatri~Sundara Padmanabhan, Subha Puttagunta, Kalpesh Krishna, Leslie Baker, Norbert Kalb, Vamsi Bedapudi, Adam Kurzrok, Shuntong Lei, Anthony Yu, Oren Litvin, Xiang Zhou, Zhichun Wu, Sam Sobell, Andrea Siciliano, Alan Papir, Robby Neale, Jonas Bragagnolo, Tej Toor, Tina Chen, Valentin Anklin, Feiran Wang, Richie Feng, Milad
  Gholami, Kevin Ling, Lijuan Liu, Jules Walter, Hamid Moghaddam, Arun Kishore, Jakub Adamek, Tyler Mercado, Jonathan Mallinson, Siddhinita Wandekar, Stephen Cagle, Eran Ofek, Guillermo Garrido, Clemens Lombriser, Maksim Mukha, Botu Sun, Hafeezul~Rahman Mohammad, Josip Matak, Yadi Qian, Vikas Peswani, Pawel Janus, Quan Yuan, Leif Schelin, Oana David, Ankur Garg, Yifan He, Oleksii Duzhyi, Anton Älgmyr, Timothée Lottaz, Qi~Li, Vikas Yadav, Luyao Xu, Alex Chinien, Rakesh Shivanna, Aleksandr Chuklin, Josie Li, Carrie Spadine, Travis Wolfe, Kareem Mohamed, Subhabrata Das, Zihang Dai, Kyle He, Daniel von Dincklage, Shyam Upadhyay, Akanksha Maurya, Luyan Chi, Sebastian Krause, Khalid Salama, Pam~G Rabinovitch, Pavan Kumar~Reddy M, Aarush Selvan, Mikhail Dektiarev, Golnaz Ghiasi, Erdem Guven, Himanshu Gupta, Boyi Liu, Deepak Sharma, Idan~Heimlich Shtacher, Shachi Paul, Oscar Akerlund, François-Xavier Aubet, Terry Huang, Chen Zhu, Eric Zhu, Elico Teixeira, Matthew Fritze, Francesco Bertolini, Liana-Eleonora
  Marinescu, Martin Bölle, Dominik Paulus, Khyatti Gupta, Tejasi Latkar, Max Chang, Jason Sanders, Roopa Wilson, Xuewei Wu, Yi-Xuan Tan, Lam~Nguyen Thiet, Tulsee Doshi, Sid Lall, Swaroop Mishra, Wanming Chen, Thang Luong, Seth Benjamin, Jasmine Lee, Ewa Andrejczuk, Dominik Rabiej, Vipul Ranjan, Krzysztof Styrc, Pengcheng Yin, Jon Simon, Malcolm~Rose Harriott, Mudit Bansal, Alexei Robsky, Geoff Bacon, David Greene, Daniil Mirylenka, Chen Zhou, Obaid Sarvana, Abhimanyu Goyal, Samuel Andermatt, Patrick Siegler, Ben Horn, Assaf Israel, Francesco Pongetti, Chih-Wei~"Louis" Chen, Marco Selvatici, Pedro Silva, Kathie Wang, Jackson Tolins, Kelvin Guu, Roey Yogev, Xiaochen Cai, Alessandro Agostini, Maulik Shah, Hung Nguyen, Noah~Ó Donnaile, Sébastien Pereira, Linda Friso, Adam Stambler, Adam Kurzrok, Chenkai Kuang, Yan Romanikhin, Mark Geller, ZJ~Yan, Kane Jang, Cheng-Chun Lee, Wojciech Fica, Eric Malmi, Qijun Tan, Dan Banica, Daniel Balle, Ryan Pham, Yanping Huang, Diana Avram, Hongzhi Shi, Jasjot Singh, Chris
  Hidey, Niharika Ahuja, Pranab Saxena, Dan Dooley, Srividya~Pranavi Potharaju, Eileen O'Neill, Anand Gokulchandran, Ryan Foley, Kai Zhao, Mike Dusenberry, Yuan Liu, Pulkit Mehta, Ragha Kotikalapudi, Chalence Safranek-Shrader, Andrew Goodman, Joshua Kessinger, Eran Globen, Prateek Kolhar, Chris Gorgolewski, Ali Ibrahim, Yang Song, Ali Eichenbaum, Thomas Brovelli, Sahitya Potluri, Preethi Lahoti, Cip Baetu, Ali Ghorbani, Charles Chen, Andy Crawford, Shalini Pal, Mukund Sridhar, Petru Gurita, Asier Mujika, Igor Petrovski, Pierre-Louis Cedoz, Chenmei Li, Shiyuan Chen, Niccolò~Dal Santo, Siddharth Goyal, Jitesh Punjabi, Karthik Kappaganthu, Chester Kwak, Pallavi LV, Sarmishta Velury, Himadri Choudhury, Jamie Hall, Premal Shah, Ricardo Figueira, Matt Thomas, Minjie Lu, Ting Zhou, Chintu Kumar, Thomas Jurdi, Sharat Chikkerur, Yenai Ma, Adams Yu, Soo Kwak, Victor Ähdel, Sujeevan Rajayogam, Travis Choma, Fei Liu, Aditya Barua, Colin Ji, Ji~Ho Park, Vincent Hellendoorn, Alex Bailey, Taylan Bilal, Huanjie Zhou,
  Mehrdad Khatir, Charles Sutton, Wojciech Rzadkowski, Fiona Macintosh, Konstantin Shagin, Paul Medina, Chen Liang, Jinjing Zhou, Pararth Shah, Yingying Bi, Attila Dankovics, Shipra Banga, Sabine Lehmann, Marissa Bredesen, Zifan Lin, John~Eric Hoffmann, Jonathan Lai, Raynald Chung, Kai Yang, Nihal Balani, Arthur Bražinskas, Andrei Sozanschi, Matthew Hayes, Héctor~Fernández Alcalde, Peter Makarov, Will Chen, Antonio Stella, Liselotte Snijders, Michael Mandl, Ante Kärrman, Paweł Nowak, Xinyi Wu, Alex Dyck, Krishnan Vaidyanathan, Raghavender R, Jessica Mallet, Mitch Rudominer, Eric Johnston, Sushil Mittal, Akhil Udathu, Janara Christensen, Vishal Verma, Zach Irving, Andreas Santucci, Gamaleldin Elsayed, Elnaz Davoodi, Marin Georgiev, Ian Tenney, Nan Hua, Geoffrey Cideron, Edouard Leurent, Mahmoud Alnahlawi, Ionut Georgescu, Nan Wei, Ivy Zheng, Dylan Scandinaro, Heinrich Jiang, Jasper Snoek, Mukund Sundararajan, Xuezhi Wang, Zack Ontiveros, Itay Karo, Jeremy Cole, Vinu Rajashekhar, Lara Tumeh, Eyal
  Ben-David, Rishub Jain, Jonathan Uesato, Romina Datta, Oskar Bunyan, Shimu Wu, John Zhang, Piotr Stanczyk, Ye~Zhang, David Steiner, Subhajit Naskar, Michael Azzam, Matthew Johnson, Adam Paszke, Chung-Cheng Chiu, Jaume~Sanchez Elias, Afroz Mohiuddin, Faizan Muhammad, Jin Miao, Andrew Lee, Nino Vieillard, Jane Park, Jiageng Zhang, Jeff Stanway, Drew Garmon, Abhijit Karmarkar, Zhe Dong, Jong Lee, Aviral Kumar, Luowei Zhou, Jonathan Evens, William Isaac, Geoffrey Irving, Edward Loper, Michael Fink, Isha Arkatkar, Nanxin Chen, Izhak Shafran, Ivan Petrychenko, Zhe Chen, Johnson Jia, Anselm Levskaya, Zhenkai Zhu, Peter Grabowski, Yu~Mao, Alberto Magni, Kaisheng Yao, Javier Snaider, Norman Casagrande, Evan Palmer, Paul Suganthan, Alfonso Castaño, Irene Giannoumis, Wooyeol Kim, Mikołaj Rybiński, Ashwin Sreevatsa, Jennifer Prendki, David Soergel, Adrian Goedeckemeyer, Willi Gierke, Mohsen Jafari, Meenu Gaba, Jeremy Wiesner, Diana~Gage Wright, Yawen Wei, Harsha Vashisht, Yana Kulizhskaya, Jay Hoover, Maigo Le,
  Lu~Li, Chimezie Iwuanyanwu, Lu~Liu, Kevin Ramirez, Andrey Khorlin, Albert Cui, Tian LIN, Marcus Wu, Ricardo Aguilar, Keith Pallo, Abhishek Chakladar, Ginger Perng, Elena~Allica Abellan, Mingyang Zhang, Ishita Dasgupta, Nate Kushman, Ivo Penchev, Alena Repina, Xihui Wu, Tom van~der Weide, Priya Ponnapalli, Caroline Kaplan, Jiri Simsa, Shuangfeng Li, Olivier Dousse, Fan Yang, Jeff Piper, Nathan Ie, Rama Pasumarthi, Nathan Lintz, Anitha Vijayakumar, Daniel Andor, Pedro Valenzuela, Minnie Lui, Cosmin Paduraru, Daiyi Peng, Katherine Lee, Shuyuan Zhang, Somer Greene, Duc~Dung Nguyen, Paula Kurylowicz, Cassidy Hardin, Lucas Dixon, Lili Janzer, Kiam Choo, Ziqiang Feng, Biao Zhang, Achintya Singhal, Dayou Du, Dan McKinnon, Natasha Antropova, Tolga Bolukbasi, Orgad Keller, David Reid, Daniel Finchelstein, Maria~Abi Raad, Remi Crocker, Peter Hawkins, Robert Dadashi, Colin Gaffney, Ken Franko, Anna Bulanova, Rémi Leblond, Shirley Chung, Harry Askham, Luis~C. Cobo, Kelvin Xu, Felix Fischer, Jun Xu, Christina Sorokin,
  Chris Alberti, Chu-Cheng Lin, Colin Evans, Alek Dimitriev, Hannah Forbes, Dylan Banarse, Zora Tung, Mark Omernick, Colton Bishop, Rachel Sterneck, Rohan Jain, Jiawei Xia, Ehsan Amid, Francesco Piccinno, Xingyu Wang, Praseem Banzal, Daniel~J. Mankowitz, Alex Polozov, Victoria Krakovna, Sasha Brown, MohammadHossein Bateni, Dennis Duan, Vlad Firoiu, Meghana Thotakuri, Tom Natan, Matthieu Geist, Ser tan Girgin, Hui Li, Jiayu Ye, Ofir Roval, Reiko Tojo, Michael Kwong, James Lee-Thorp, Christopher Yew, Danila Sinopalnikov, Sabela Ramos, John Mellor, Abhishek Sharma, Kathy Wu, David Miller, Nicolas Sonnerat, Denis Vnukov, Rory Greig, Jennifer Beattie, Emily Caveness, Libin Bai, Julian Eisenschlos, Alex Korchemniy, Tomy Tsai, Mimi Jasarevic, Weize Kong, Phuong Dao, Zeyu Zheng, Frederick Liu, Fan Yang, Rui Zhu, Tian~Huey Teh, Jason Sanmiya, Evgeny Gladchenko, Nejc Trdin, Daniel Toyama, Evan Rosen, Sasan Tavakkol, Linting Xue, Chen Elkind, Oliver Woodman, John Carpenter, George Papamakarios, Rupert Kemp, Sushant
  Kafle, Tanya Grunina, Rishika Sinha, Alice Talbert, Diane Wu, Denese Owusu-Afriyie, Cosmo Du, Chloe Thornton, Jordi Pont-Tuset, Pradyumna Narayana, Jing Li, Saaber Fatehi, John Wieting, Omar Ajmeri, Benigno Uria, Yeongil Ko, Laura Knight, Amélie Héliou, Ning Niu, Shane Gu, Chenxi Pang, Yeqing Li, Nir Levine, Ariel Stolovich, Rebeca Santamaria-Fernandez, Sonam Goenka, Wenny Yustalim, Robin Strudel, Ali Elqursh, Charlie Deck, Hyo Lee, Zonglin Li, Kyle Levin, Raphael Hoffmann, Dan Holtmann-Rice, Olivier Bachem, Sho Arora, Christy Koh, Soheil~Hassas Yeganeh, Siim Põder, Mukarram Tariq, Yanhua Sun, Lucian Ionita, Mojtaba Seyedhosseini, Pouya Tafti, Zhiyu Liu, Anmol Gulati, Jasmine Liu, Xinyu Ye, Bart Chrzaszcz, Lily Wang, Nikhil Sethi, Tianrun Li, Ben Brown, Shreya Singh, Wei Fan, Aaron Parisi, Joe Stanton, Vinod Koverkathu, Christopher~A. Choquette-Choo, Yunjie Li, TJ~Lu, Abe Ittycheriah, Prakash Shroff, Mani Varadarajan, Sanaz Bahargam, Rob Willoughby, David Gaddy, Guillaume Desjardins, Marco Cornero, Brona
  Robenek, Bhavishya Mittal, Ben Albrecht, Ashish Shenoy, Fedor Moiseev, Henrik Jacobsson, Alireza Ghaffarkhah, Morgane Rivière, Alanna Walton, Clément Crepy, Alicia Parrish, Zongwei Zhou, Clement Farabet, Carey Radebaugh, Praveen Srinivasan, Claudia van~der Salm, Andreas Fidjeland, Salvatore Scellato, Eri Latorre-Chimoto, Hanna Klimczak-Plucińska, David Bridson, Dario de~Cesare, Tom Hudson, Piermaria Mendolicchio, Lexi Walker, Alex Morris, Matthew Mauger, Alexey Guseynov, Alison Reid, Seth Odoom, Lucia Loher, Victor Cotruta, Madhavi Yenugula, Dominik Grewe, Anastasia Petrushkina, Tom Duerig, Antonio Sanchez, Steve Yadlowsky, Amy Shen, Amir Globerson, Lynette Webb, Sahil Dua, Dong Li, Surya Bhupatiraju, Dan Hurt, Haroon Qureshi, Ananth Agarwal, Tomer Shani, Matan Eyal, Anuj Khare, Shreyas~Rammohan Belle, Lei Wang, Chetan Tekur, Mihir~Sanjay Kale, Jinliang Wei, Ruoxin Sang, Brennan Saeta, Tyler Liechty, Yi~Sun, Yao Zhao, Stephan Lee, Pandu Nayak, Doug Fritz, Manish~Reddy Vuyyuru, John Aslanides, Nidhi Vyas,
  Martin Wicke, Xiao Ma, Evgenii Eltyshev, Nina Martin, Hardie Cate, James Manyika, Keyvan Amiri, Yelin Kim, Xi~Xiong, Kai Kang, Florian Luisier, Nilesh Tripuraneni, David Madras, Mandy Guo, Austin Waters, Oliver Wang, Joshua Ainslie, Jason Baldridge, Han Zhang, Garima Pruthi, Jakob Bauer, Feng Yang, Riham Mansour, Jason Gelman, Yang Xu, George Polovets, Ji~Liu, Honglong Cai, Warren Chen, XiangHai Sheng, Emily Xue, Sherjil Ozair, Christof Angermueller, Xiaowei Li, Anoop Sinha, Weiren Wang, Julia Wiesinger, Emmanouil Koukoumidis, Yuan Tian, Anand Iyer, Madhu Gurumurthy, Mark Goldenson, Parashar Shah, MK~Blake, Hongkun Yu, Anthony Urbanowicz, Jennimaria Palomaki, Chrisantha Fernando, Ken Durden, Harsh Mehta, Nikola Momchev, Elahe Rahimtoroghi, Maria Georgaki, Amit Raul, Sebastian Ruder, Morgan Redshaw, Jinhyuk Lee, Denny Zhou, Komal Jalan, Dinghua Li, Blake Hechtman, Parker Schuh, Milad Nasr, Kieran Milan, Vladimir Mikulik, Juliana Franco, Tim Green, Nam Nguyen, Joe Kelley, Aroma Mahendru, Andrea Hu, Joshua
  Howland, Ben Vargas, Jeffrey Hui, Kshitij Bansal, Vikram Rao, Rakesh Ghiya, Emma Wang, Ke~Ye, Jean~Michel Sarr, Melanie~Moranski Preston, Madeleine Elish, Steve Li, Aakash Kaku, Jigar Gupta, Ice Pasupat, Da-Cheng Juan, Milan Someswar, Tejvi M., Xinyun Chen, Aida Amini, Alex Fabrikant, Eric Chu, Xuanyi Dong, Amruta Muthal, Senaka Buthpitiya, Sarthak Jauhari, Nan Hua, Urvashi Khandelwal, Ayal Hitron, Jie Ren, Larissa Rinaldi, Shahar Drath, Avigail Dabush, Nan-Jiang Jiang, Harshal Godhia, Uli Sachs, Anthony Chen, Yicheng Fan, Hagai Taitelbaum, Hila Noga, Zhuyun Dai, James Wang, Chen Liang, Jenny Hamer, Chun-Sung Ferng, Chenel Elkind, Aviel Atias, Paulina Lee, Vít Listík, Mathias Carlen, Jan van~de Kerkhof, Marcin Pikus, Krunoslav Zaher, Paul Müller, Sasha Zykova, Richard Stefanec, Vitaly Gatsko, Christoph Hirnschall, Ashwin Sethi, Xingyu~Federico Xu, Chetan Ahuja, Beth Tsai, Anca Stefanoiu, Bo~Feng, Keshav Dhandhania, Manish Katyal, Akshay Gupta, Atharva Parulekar, Divya Pitta, Jing Zhao, Vivaan Bhatia,
  Yashodha Bhavnani, Omar Alhadlaq, Xiaolin Li, Peter Danenberg, Dennis Tu, Alex Pine, Vera Filippova, Abhipso Ghosh, Ben Limonchik, Bhargava Urala, Chaitanya~Krishna Lanka, Derik Clive, Yi~Sun, Edward Li, Hao Wu, Kevin Hongtongsak, Ianna Li, Kalind Thakkar, Kuanysh Omarov, Kushal Majmundar, Michael Alverson, Michael Kucharski, Mohak Patel, Mudit Jain, Maksim Zabelin, Paolo Pelagatti, Rohan Kohli, Saurabh Kumar, Joseph Kim, Swetha Sankar, Vineet Shah, Lakshmi Ramachandruni, Xiangkai Zeng, Ben Bariach, Laura Weidinger, Tu~Vu, Alek Andreev, Antoine He, Kevin Hui, Sheleem Kashem, Amar Subramanya, Sissie Hsiao, Demis Hassabis, Koray Kavukcuoglu, Adam Sadovsky, Quoc Le, Trevor Strohman, Yonghui Wu, Slav Petrov, Jeffrey Dean, and Oriol Vinyals.
\newblock Gemini: A family of highly capable multimodal models, 2024.

\bibitem{UnityDocs2023}
{Unity Technologies}.
\newblock Inner workings of the navigation system, 2023.
\newblock Accessed: 2024-06-03.

\bibitem{von2021informed}
Laura Von~Rueden, Sebastian Mayer, Katharina Beckh, Bogdan Georgiev, Sven Giesselbach, Raoul Heese, Birgit Kirsch, Julius Pfrommer, Annika Pick, Rajkumar Ramamurthy, et~al.
\newblock Informed machine learning--a taxonomy and survey of integrating prior knowledge into learning systems.
\newblock {\em IEEE Transactions on Knowledge and Data Engineering}, 35(1):614--633, 2021.

\bibitem{APFOG}
C.~Warren.
\newblock Global path planning using artificial potential fields.
\newblock In {\em 1989 IEEE International Conference on Robotics and Automation}, pages 316,317,318,319,320,321, Los Alamitos, CA, USA, may 1989. IEEE Computer Society.

\bibitem{oldAPF}
C.W. Warren.
\newblock Multiple robot path coordination using artificial potential fields.
\newblock In {\em Proceedings., IEEE International Conference on Robotics and Automation}, pages 500--505 vol.1, 1990.

\bibitem{MDPILIDar}
Pan Wei, Lucas Cagle, Tasmia Reza, John Ball, and James Gafford.
\newblock Lidar and camera detection fusion in a real-time industrial multi-sensor collision avoidance system.
\newblock {\em Electronics}, 7(6), 2018.

\bibitem{TWIN}
Wang Wenna, Ding Weili, Hua Changchun, Zhang Heng, Feng Haibing, and Yao Yao.
\newblock A digital twin for 3d path planning of large-span curved-arm gantry robot.
\newblock {\em Robotics and Computer-Integrated Manufacturing}, 76:102330, 2022.

\bibitem{wu2022apf}
Daohua Wu, Lisheng Wei, Guanling Wang, Li~Tian, and Guangzhen Dai.
\newblock Apf-irrt*: An improved informed rapidly-exploring random trees-star algorithm by introducing artificial potential field method for mobile robot path planning.
\newblock {\em Applied Sciences}, 12(21):10905, 2022.

\bibitem{wu2023robot}
Zhengtian Wu, Jinyu Dai, Baoping Jiang, and Hamid~Reza Karimi.
\newblock Robot path planning based on artificial potential field with deterministic annealing.
\newblock {\em ISA transactions}, 138:74--87, 2023.

\bibitem{gpt35t}
Junjie Ye, Xuanting Chen, Nuo Xu, Can Zu, Zekai Shao, Shichun Liu, Yuhan Cui, Zeyang Zhou, Chao Gong, Yang Shen, Jie Zhou, Siming Chen, Tao Gui, Qi~Zhang, and Xuanjing Huang.
\newblock A comprehensive capability analysis of gpt-3 and gpt-3.5 series models, 2023.

\bibitem{EricTrustGPT}
Yang Ye, Hengxu You, and Jing Du.
\newblock Improved trust in human-robot collaboration with chatgpt.
\newblock {\em IEEE Access}, 11:55748--55754, 2023.

\bibitem{yilmaz2016indoor}
Alper Yilmaz and Ashish Gupta.
\newblock Indoor positioning using visual and inertial sensors.
\newblock In {\em 2016 IEEE SENSORS}, pages 1--3. IEEE, 2016.

\bibitem{BIMLocalationzerpoint2}
Huan Yin, Jia~Min Liew, Wai~Leong Lee, Marcelo~H. Ang, Ker-Wei Yeoh, and Justin.
\newblock Towards bim-based robot localization: a real-world case study.
\newblock In Thomas Linner, Borja García~de Soto, Rongbo Hu, Ioannis Brilakis, Thomas Bock, Wen Pan, Alessandro Carbonari, Daniel Castro, Harrison Mesa, Chen Feng, Martin Fischer, Cynthia Brosque, Vicente Gonzalez, Daniel Hall, Ming~Shan Ng, Vineet Kamat, Ci-Jyun Liang, Zoubeir Lafhaj, Wei Pan, Mi~Pan, and Zhenhua Zhu, editors, {\em Proceedings of the 39th International Symposium on Automation and Robotics in Construction}, pages 71--77, Bogot�, Colombia, July 2022. International Association for Automation and Robotics in Construction (IAARC).

\bibitem{zabin2022applications}
Asem Zabin, Vicente~A Gonz{\'a}lez, Yang Zou, and Robert Amor.
\newblock Applications of machine learning to bim: A systematic literature review.
\newblock {\em Advanced Engineering Informatics}, 51:101474, 2022.

\bibitem{zaitsev2017generalized}
Dmitry~A Zaitsev.
\newblock A generalized neighborhood for cellular automata.
\newblock {\em Theoretical Computer Science}, 666:21--35, 2017.

\bibitem{zhang2024agv}
Dengxing Zhang, Chen Chen, and Guanyu Zhang.
\newblock Agv path planning based on improved a-star algorithm.
\newblock In {\em 2024 IEEE 7th Advanced Information Technology, Electronic and Automation Control Conference (IAEAC)}, volume~7, pages 1590--1595. IEEE, 2024.

\bibitem{zhang2022towards}
Jiale Zhang, Hanbin Luo, and Jie Xu.
\newblock Towards fully bim-enabled building automation and robotics: A perspective of lifecycle information flow.
\newblock {\em Computers in Industry}, 135:103570, 2022.

\bibitem{zhou2020accurate}
Xiaoping Zhou, Qingsheng Xie, Maozu Guo, Jichao Zhao, and Jia Wang.
\newblock Accurate and efficient indoor pathfinding based on building information modeling data.
\newblock {\em IEEE Transactions on Industrial Informatics}, 16(12):7459--7468, 2020.

\bibitem{Videogame}
Moh. Zikky.
\newblock Review of a* (a star) navigation mesh pathfinding as the alternative of artificial intelligent for ghosts agent on the pacman game.
\newblock {\em EMITTER International Journal of Engineering Technology}, 4(1):141--149, Aug. 2016.

\end{thebibliography}

%





\begin{IEEEbiography}[{\includegraphics[width=1in,height=1.25in,clip,keepaspectratio]{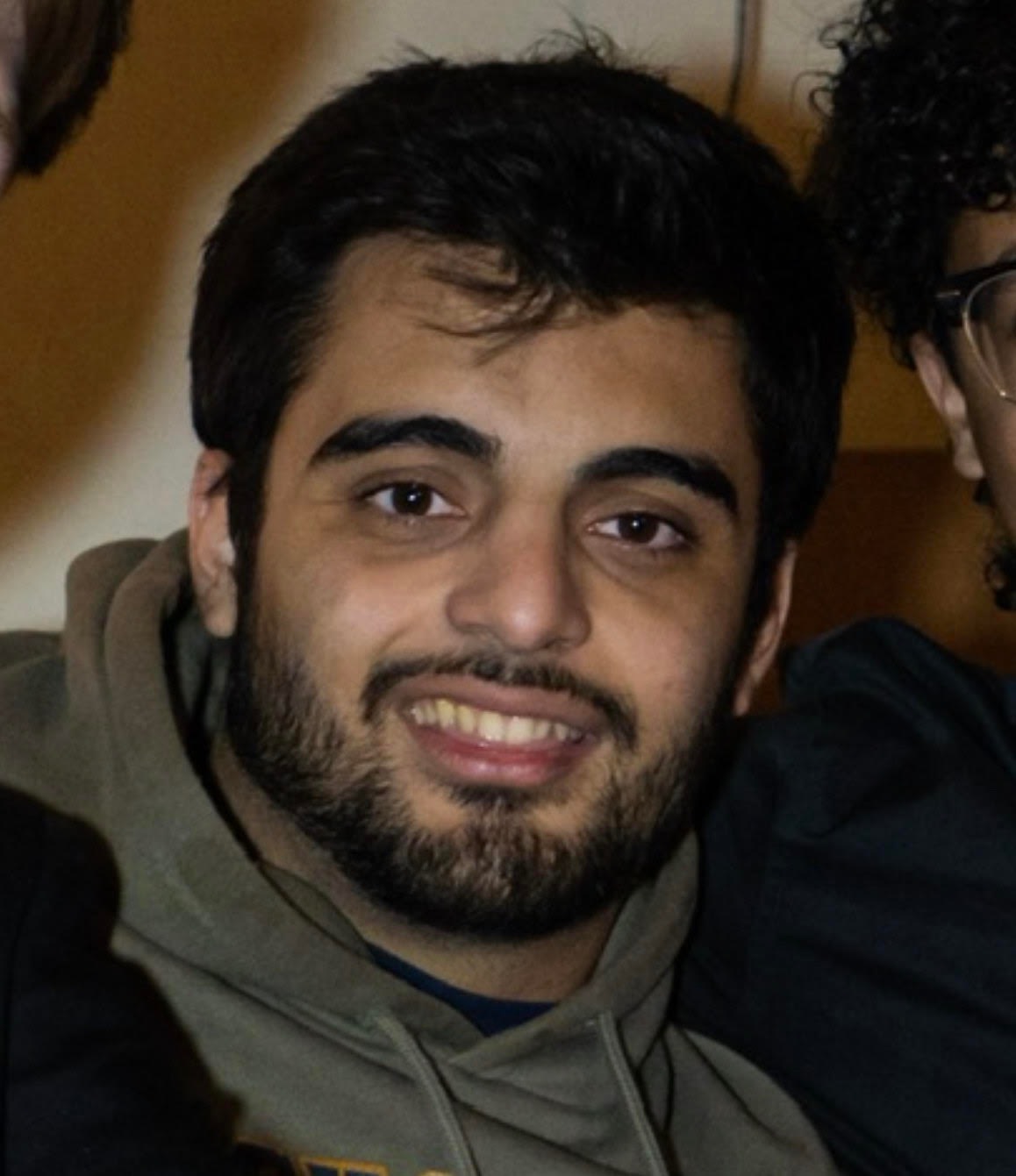}}]{Mani Amani} received the B.S. in Cognitive Science from the University of California, San Diego. He is currently a Ph.D student at San Diego State University and University of California, San Diego.
\end{IEEEbiography}

\begin{IEEEbiography}
[{\includegraphics[width=1in,height=1.25in,clip,keepaspectratio]{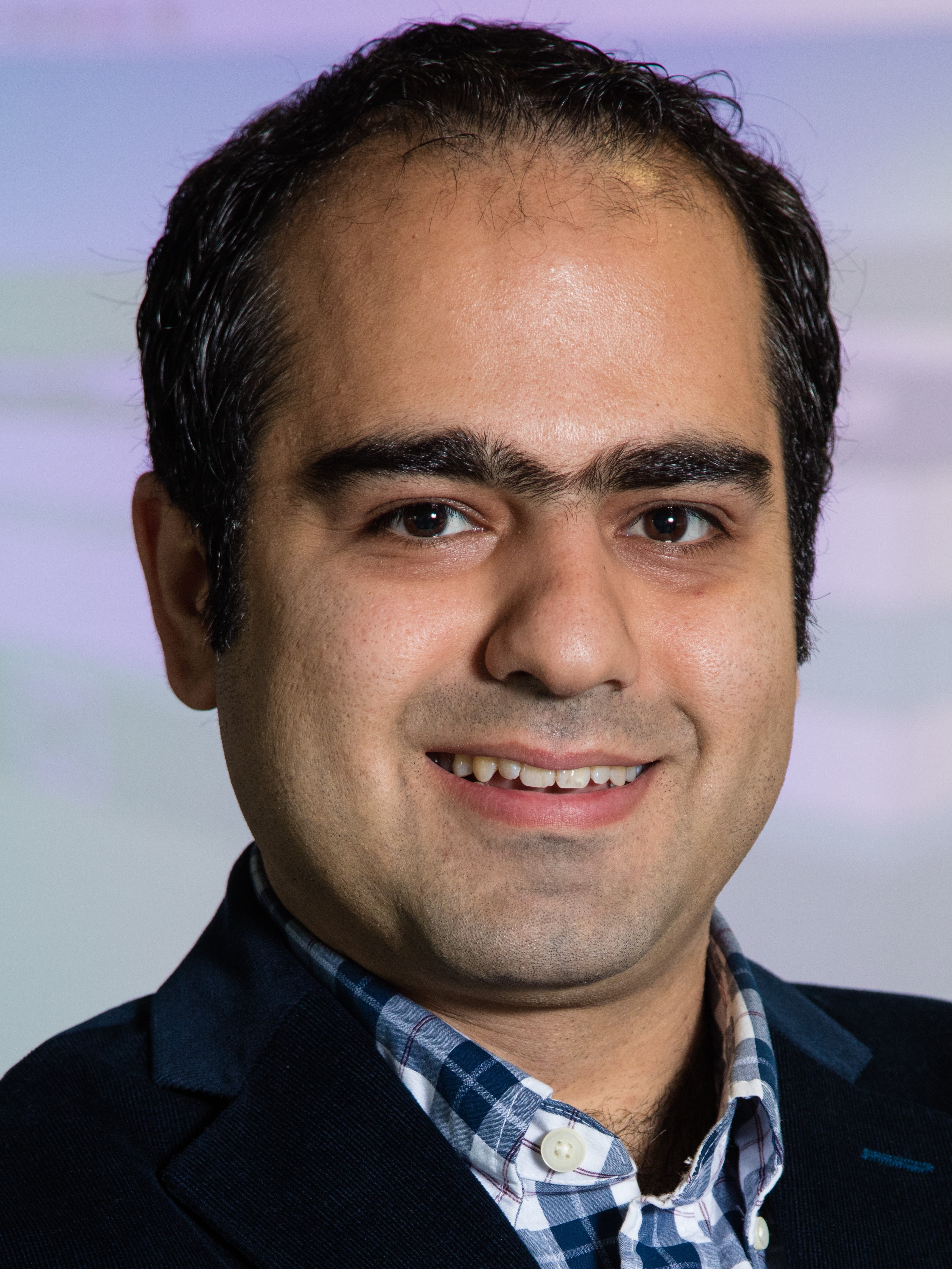}}]{Reza Akhavian} received a B.S. (2010) in Civil Engineering from the University of Tehran, and a M.S. (2012) and a Ph.D. (2015) in Civil Engineering from the University of Central Florida. He is currently an Associate Professor at the Department of Civil, Construction, and Environmental Engineering, and the director of the Data-informed Construction Engineering (DiCE) research lab at San Diego State University. Dr. Akhavian has received several honors and awards including the NSF CAREER Award. His research focuses on advancing the theories and adoption of technologies such as robotics, artificial intelligence (AI), internet-of-things (IoT), cyber-physical systems (CPS), and augmented and virtual reality (AR/VR) in engineering and construction domains. He is the elected Secretary of the American Society of Civil Engineers (ASCE) Visualization, Information Modeling and Simulation (VIMS) Committee, and serves on the editorial board of the ASCE Journal of Construction Engineering and Management and Elsevier Advanced Engineering Informatics.

\end{IEEEbiography}


\EOD

\end{document}